%% file: main.tex
\documentclass{article}

\PassOptionsToPackage{numbers}{natbib}

\usepackage[final]{neurips_2022}

\usepackage[utf8]{inputenc} 
\usepackage[T1]{fontenc}    
\usepackage{hyperref}       
\usepackage{url}            
\usepackage{booktabs}       
\usepackage{amsfonts}       
\usepackage{nicefrac}       
\usepackage{microtype}      
\usepackage{xcolor}         

\usepackage{amsmath}
\usepackage{amssymb}
\usepackage{mathtools}
\usepackage{amsthm}

\input{xcl_math}

\theoremstyle{plain}
\theoremstyle{definition}

\theoremstyle{remark}

\usepackage{times}
\usepackage{helvet}
\usepackage{courier}
\usepackage{caption}
\usepackage{color}
\usepackage{epsfig}
\usepackage{bm}
\usepackage{makecell}
\usepackage{multirow}
\usepackage{algorithm}
\usepackage{algorithmic}
\usepackage{adjustbox}
\usepackage{bbm}
\usepackage{epstopdf}
\usepackage{threeparttable}
\usepackage{tablefootnote}
\usepackage[normalem]{ulem}
\usepackage{color,soul}

\usepackage[subtle]{savetrees}
\usepackage{comment}
\usepackage{subcaption}

\DeclareMathAlphabet{\mathcal}{OMS}{cmsy}{m}{n}

\newcommand{\vpi}{\boldsymbol{\pi}}

\newcommand{\vxi}{\boldsymbol{\xi}}
\newcommand{\vnu}{\boldsymbol{\nu}}

\title{ZooD: Exploiting Model Zoo for \\ Out-of-Distribution Generalization}

\author{%
  Qishi Dong \textsuperscript{\rm 2,1}\thanks{Equal Contribution. This work was carried out at Huawei Noah's Ark Lab.},  
  Awais Muhammad \textsuperscript{\rm 3,1}\footnotemark[1]\, ,
  Fengwei Zhou \textsuperscript{\rm 1}\footnotemark[1]\, , 
  Chuanlong Xie \textsuperscript{\rm 4,1}\thanks{Correspondence to: li.zhenguo@huawei.com;  clxie@bnu.edu.cn}\, , 
  Tianyang Hu \textsuperscript{\rm 1}, \\
  \textbf{Yongxin Yang \textsuperscript{\rm 1}, 
  Sung-Ho Bae \textsuperscript{\rm 3}, 
  Zhenguo Li \textsuperscript{\rm 1}\footnotemark[2]}
  \\
  ~\\
  \textsuperscript{\rm 1} Huawei Noah's Ark Lab, \\
  \textsuperscript{\rm 2} Hong Kong Baptist University, \\
  \textsuperscript{\rm 3} Kyung-Hee University, \\
  \textsuperscript{\rm 4} Beijing Normal University \\
}

\begin{document}

\maketitle

\begin{abstract}

Recent advances on large-scale pre-training have shown great potentials of leveraging a large set of Pre-Trained Models (PTMs) for improving Out-of-Distribution (OoD) generalization, for which the goal is to perform well on possible unseen domains after fine-tuning on multiple training domains. However, maximally exploiting a zoo of PTMs is challenging since fine-tuning all possible combinations of PTMs is computationally prohibitive while accurate selection of PTMs requires tackling the possible data distribution shift for OoD tasks. In this work, we propose ZooD, a paradigm for PTMs ranking and ensemble with feature selection. Our proposed metric ranks PTMs by quantifying inter-class discriminability and inter-domain stability of the features extracted by the PTMs in a leave-one-domain-out cross-validation manner. The top-K ranked models are then aggregated for the target OoD task. To avoid accumulating noise induced by model ensemble, we propose an efficient variational EM algorithm to select informative features. We evaluate our paradigm on a diverse model zoo consisting of 35 models for various OoD tasks and demonstrate: (i) model ranking is better correlated with fine-tuning ranking than previous methods and up to 9859x faster than brute-force fine-tuning; (ii) OoD generalization after model ensemble with feature selection outperforms the state-of-the-art methods and the accuracy on most challenging task DomainNet is improved from 46.5\% to 50.6\%. Furthermore, we provide the fine-tuning results of 35 PTMs on 7 OoD datasets, hoping to help the research of model zoo and OoD generalization. Code will be available at \href{https://gitee.com/mindspore/models/tree/master/research/cv/zood}{https://gitee.com/mindspore/models/tree/master/research/cv/zood}.
\end{abstract}

\section{Introduction} 

Although data Independent and Identically Distributed (IID) is a primary assumption behind most machine learning systems, it does not hold in many real-world scenarios due to continuous distribution shifts~\cite{koh2021wilds,ye2022ood}. Machine learning models encounter serious performance degradation~\cite{barbu2019objectnet,hendrycks2019robustness,hendrycks2021nae} in such Out-of-Distribution (OoD) scenarios. To alleviate the accuracy degradation caused by distribution shifts, numerous algorithms have been proposed~\cite{arjovsky2019invariant,ahuja2020invariant,krueger2021out,Li2018LearningTG,bai2020decaug,Kuang2018,shen2019stable,he2020towards,dapello2020simulating,li2020network,bai2021ood}.
Recently, \citet{gulrajani2020search} have argued for the systematic comparisons of OoD algorithms and introduced a standard and rigorous test bed called DomainBed. Their experimental comparison has raised concerns about the effectiveness of OoD algorithms since they often fail to outperform the simple Empirical Risk Minimization (ERM).

On the other hand, recent works~\cite{Hendrycks2020PretrainedTI, Albuquerque2020ImprovingOG, Yi2021ImprovedOG, Radford2021LearningTV} have shown the advantages of pre-training for improving OoD generalization, i.e., learning from multiple training domains in order to generalize to an unseen domain. The availability of a large set of Pre-Trained Models (PTMs) provides a huge potential for solving various OoD tasks.
However, it is challenging to sufficiently exploit the power of a model zoo (a large set of PTMs).
One naive approach could be fine-tuning all possible combinations of PTMs on the target dataset and choosing the best-performing one, which is computationally expensive especially when the number of PTMs and the data size are large. Besides, fine-tuning may also require exhaustive hyper-parameter search and encounter the risk of over-fitting~\cite{you2021logme}.  

\begin{figure*}[t]
    \centering
    \includegraphics[width=0.95\linewidth]{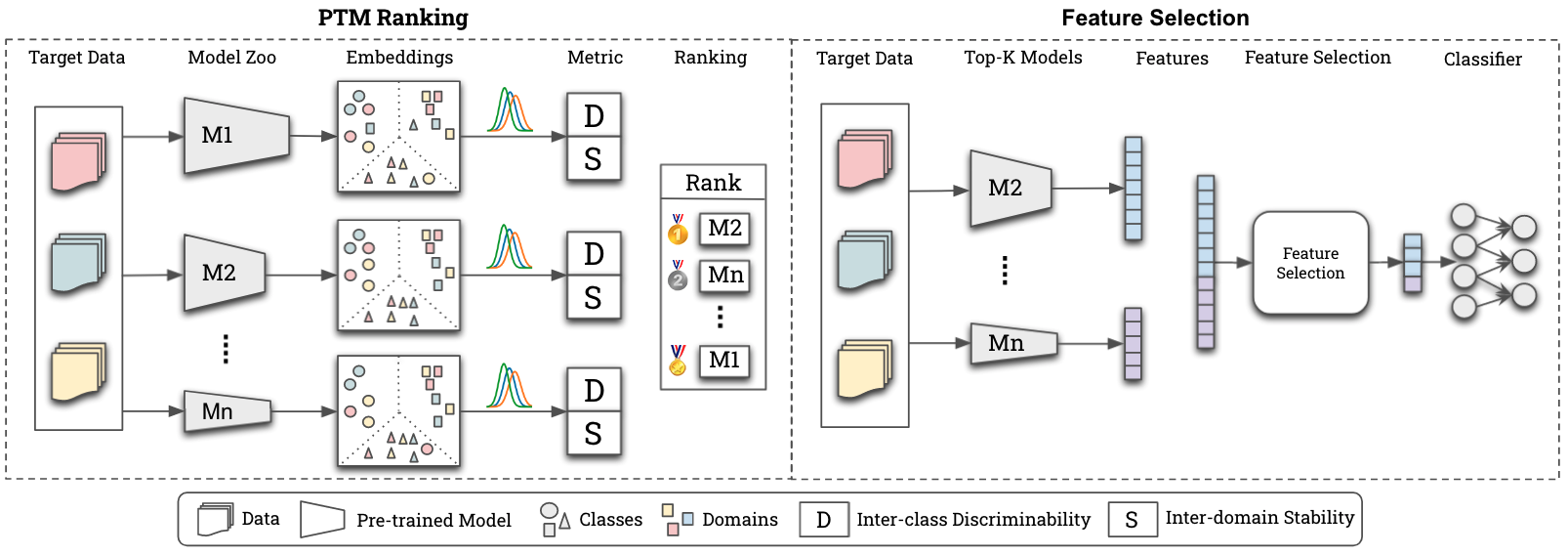}
    \caption{An overview of ZooD. Given a task with multiple training domains, the model ranking component evaluates and selects the top-K models that generalize well on this task. The features from selected models are then aggregated and denoised based on the feature selection component.}
    \label{fig:overview}
    \vspace{-15pt}
\end{figure*}
Recently, many ranking metrics have been proposed to estimate the transferability of models under IID assumption~\cite{Bao2019AnIA, Tran2019TransferabilityAH, Nguyen2020LEEPAN, you2021logme, you2021ranking}. However, ranking a zoo of models for generalization on unseen distribution shifts is more challenging compared with the IID setting. 
Moreover, even if a metric can correctly evaluate the transferability of each PTM, simply using the best model will not fully utilize rich knowledge present in a zoo of models. But the problem is even more serious that the most transferable model will include some noise, because noise and invariant features are undistinguishable in the sense that they are all stable across domains. Previous study~\cite{ye2021towards} also pointed this out and emphasized the necessity of feature denoising. Therefore, if we leverage the model zoo by assembling relatively transferable models, the accumulation of noise features may increase memory use and hurt the predictive performance.

To solve the aforementioned problems, we propose ZooD, a paradigm to rank and aggregate a \textbf{Zoo} of PTMs for \textbf{OoD} generalization. 
An overview of our method is shown in Figure~\ref{fig:overview} . 
Given a classification task with multiple training domains, to evaluate the generalization capability of each model, we quantify both the inter-class discriminability and inter-domain stability of the features extracted from each PTM in a leave-one-domain-out cross-validation manner, i.e., choosing one domain as the validation domain and each domain rotating as the validation domain,
which is critical for identifying models that can extract domain-invariant features. Each PTM in the zoo is ranked by this quantification.
ZooD then continues with model aggregation consisting of model ensemble and feature selection. By introducing latent masks over candidate features, an efficient EM algorithm is proposed to select informative features. To tackle the intractability of the posterior, variational approximation to the true posterior using a factorizable distribution is derived. We further extend it to large-scale datasets by building a local estimator under the stochastic approximation~\cite{robbins1951stochastic}.

To demonstrate the efficacy of our method, we have performed extensive experiments with 35 diverse PTMs and 7 OoD datasets. First, we show that our ranking metric is strongly correlated with the fine-tuning performance of PTMs compared with existing IID metrics. 
Second, we illustrate the outstanding performance of ZooD on OoD datasets. 
For instance, on Office-Home, we get 85.1\% average accuracy compared with the previous SOTA of 70.6\%. Lastly, we show the speedup of our method compared with brute-force fine-tuning. ZooD gives a maximum speedup of $\approx 10000\times$ (0.27 GPU hours vs 2662.27 GPU hours), making it practical and scalable.

Finally, to speed up research and make our work more reproducible, we have devised a test bench consisting of extracted features, fine-tuning accuracy results, and ranking scores for all 35 PTMs in our model zoo. This testbed can help future research as the process of getting fine-tuning accuracy results based on  DomainBed~\cite{gulrajani2020search} for a zoo of models is computationally expensive. For instance, fine-tuning 35 models on all 7 OoD datasets costs approximately 35140 GPU hours (equivalent to 1464 GPU days or 4 GPU years). 
Concisely, our contributions are as follows:
\begin{itemize}
    \item[$\bullet$]  We propose an efficient and scalable ranking metric to gauge the generalization-ability of PTMs for unseen domains.
    \item[$\bullet$]  Using EM, we propose a method for selecting informative features and discarding invariant but noisy features in an ensemble of models.
    \item[$\bullet$] We have established a test bed for PTMs on 7 OoD datasets, including features extracted by 35 PTMs in our model zoo, fine-tuning accuracy results, and model ranking scores by different methods. 
\end{itemize}

\section{Related Work}
\textbf{Pre-training for OoD generalization.} To tackle the problem of distribution shifts between training and test data, various OoD methods~\cite{arjovsky2019invariant,ahuja2020invariant,krueger2021out,Li2018LearningTG,dou2019domain,carlucci2019domain,bai2020decaug,Kuang2018,shen2019stable,dapello2020simulating,li2020network,bai2021ood} have been proposed with the aim to learn invariant representations across different environments. 
However, a standard evaluation~\cite{gulrajani2020search} of many OoD algorithms shows that they do not significantly outperform simple ERM. 
On the other hands, recent works have shown the effectiveness of pre-trained models for OoD generalization. \citet{Yi2021ImprovedOG} theoretically showed that adversarially pre-trained models also perform better for OoD generalization. \citet{yu2021empirical} 
showed that the right choice of pre-trained models can achieve SOTA results. They also showed that IID performance is not a good indicator of OoD performance and emphasized on the importance of model selection. \citet{Albuquerque2020ImprovingOG} showed the importance of feature extractor by proposing a new OoD-based pretext task for self-supervised pre-training. 
CLIP~\cite{Radford2021LearningTV} demonstrated that large-scale pre-training on a dataset of image-text pairs results in much more robust models for downstream tasks with various distribution shifts.
Our work is based on these observations and we aim to facilitate utilization of PTMs by proposing an efficient metric as well as efficient feature ensemble and selection method.

\textbf{Ranking pre-trained models by metric design.} 
Recently, a number of metrics have been introduced to estimate transferability of source-task-learned representations for target task under IID conditions. H-score~\cite{Bao2019AnIA} estimates the transferability by finding the relationship between extracted features and target class labels. NCE~\cite{Tran2019TransferabilityAH} proposes to estimate transferability via measuring conditional entropy between source and target labels. 
LEEP~\cite{Nguyen2020LEEPAN} simplifies NCE by using the joint distribution of source and target labels to estimate log expected empirical prediction. LogME~\cite{you2021logme, you2021ranking} estimates the maximum value of label evidence given features from pre-trained models. 
These transferability metrics focus on determining the compatibility of source-task-learned representations for the target task. We, on the other hand, aim to compute stability of these features across domains in addition to source-target transferability.

\textbf{Ensemble and feature selection.} Early works have shown that model ensemble can significantly improve predictive performance~\cite{dietterich2000ensemble}. In the age of deep learning, \citet{lakshminarayanan2016simple} propose deep ensemble to measure predictive uncertainty. Similar works~\cite{osband2016deep,pawlowski2017efficient} on uncertainty estimation focus on the context of outlier detection and reinforcement learning. When facing a zoo of PTMs, it's natural to leverage the rich knowledge by assembling multiple PTMs. In prior works, 
\citet{liu2019knowledge} propose using PTMs as teacher models that distill knowledge to a target model for downstream tasks. \citet{shu2021zoo} propose Zoo-Tuning that learns to aggregate the parameters of multiple PTMs to a target model.
However, these methods require the target model must have the identical architecture as the PTMs, thus sacrificing flexibility. 

Our proposed paradigm involves selecting informative features from assembled feature extractors. In the framework of Bayesian variable selection, it is common practice to identify promising features by estimating the posterior probabilities over all potential feature subsets. Here we mainly focus on Stochastic Search Variable Selection (SSVS)~\cite{o2009review} that involves specifying priors over regression coefficients such that higher posterior probabilities will be allocated to coefficients substantially different from zero. Then the features whose coefficients have higher posterior probabilities will be selected. \citet{george1995stochastic} first propose SSVS for the linear model and conduct the posterior inference using Gibbs sampling \cite{neal2011mcmc}.
\citet{li2010bayesian} consider SSVS for regression modeling in high-dimensional spaces incorporating structural information. \citet{rovckova2014emvs} propose EMVS for efficient SSVS in high-dimentional cases with sparse estimations of posterior probabilities. Note that all aforementioned feature selection methods have inherent assumptions that observed datasets must be IID, which makes these methods difficult to use in our scenarios.

\section{ZooD for OoD Generalization}\label{sec3}

\subsection{Model Ranking}\label{sec31}

\begin{wrapfigure}{r}{0.4\textwidth}
    \centering
    \includegraphics[width=0.4\textwidth]{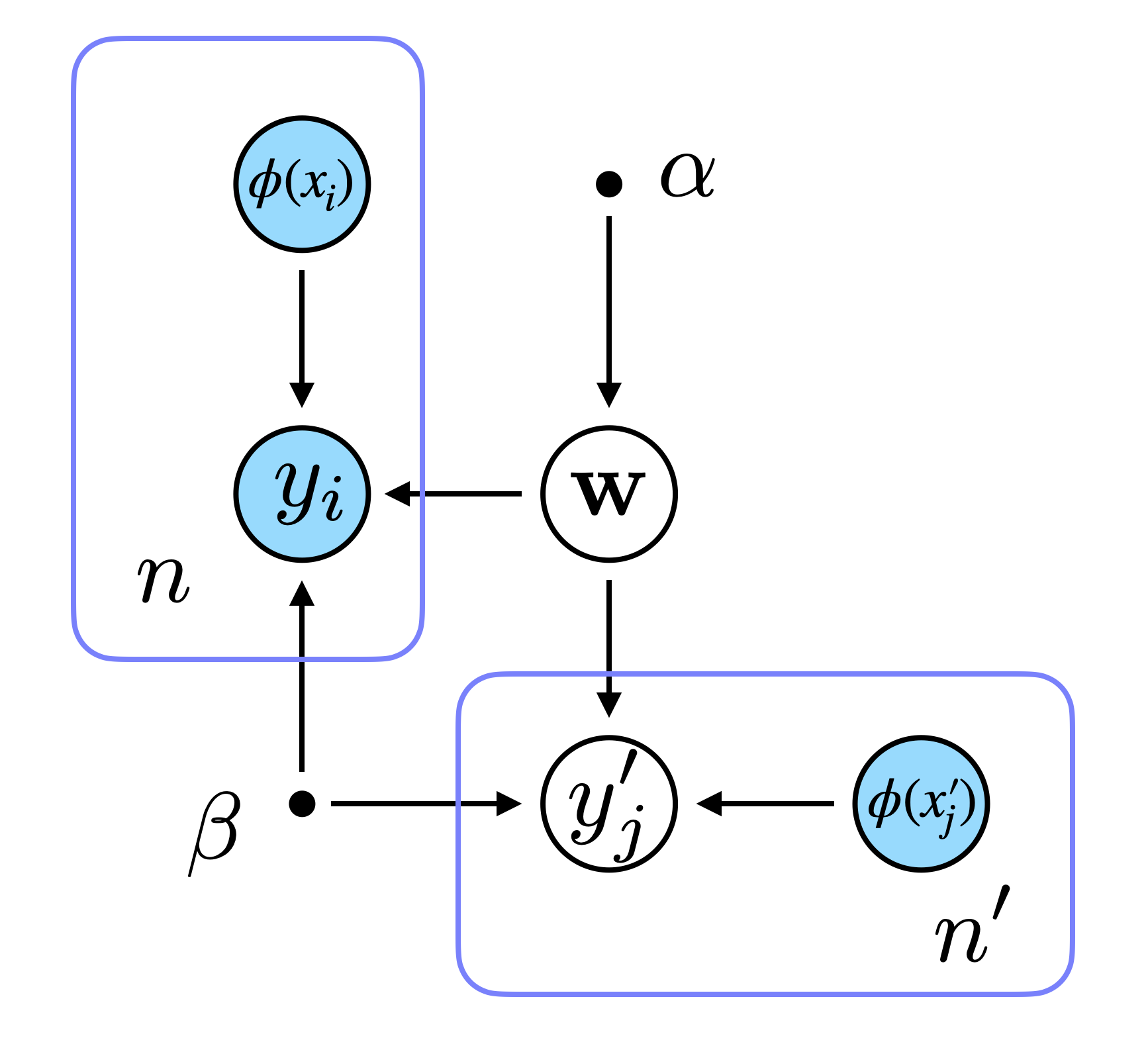}
    \caption{A directed graphical model that represents the model assumptions in (\ref{linearmodel1}). Here $\alpha$ and $\beta$ are hyper-parameters. The goal is to inference the conditional distribution of $y'_j$ given $\phi(x'_j)$, $y_i$, and $\phi(x_i)$.}
    \label{fig:model1}
    \vspace{-15pt}
\end{wrapfigure}

Assume that we have a domain distribution $\mathcal{D}$ 
from which we observe $m$ domains:
$
\big\{\mathcal{D}_1, \mathcal{D}_2, \cdots, \mathcal{D}_m \big\}. 
$
Each domain $\cD_i$ is a set of label and datum pairs, i.e. $\cD_i=\big\{(y_{ij},x_{ij}), 1\leq j \leq n_i\big\}.$
Meanwhile, we have a zoo of pre-trained feature extractors: $\mathcal{M} = \{\phi_1, \phi_2, \cdots, \phi_k, \cdots \}$.
Our objective is to select a feature extractor from $\mathcal{M}$ (e.g., $\phi_k$), such that when we train a predictor $f$ on top of it, the composed model $f \circ \phi_k$ can have the best performance on both the $m$ observed domains and unseen domains from $\mathcal{D}$.

In this section, we propose a method that facilitates model selection \emph{without} carrying out the fine-tuning step. For every model in $\mathcal{M}$, our method produces an associated score, by which we can \emph{rank} the models, such that the higher-ranked ones have a better chance to deliver stronger results after fine-tuning.

The proposed method is a combination of 1) a model transferability metric and 2) a leave-one-domain-out cross-validation scheme. More specifically, we evaluate each feature extractor $m$ times, and each time we treat the data from the held-out domain as validation data $\{(y'_j, x'_j)\}_{j=1}^{n'}$, while aggregating all remaining $(m-1)$ domains' data as the training data $\{(y_i, x_i)\}_{i=1}^n$. 
In the end, we average the $m$ values of the model transferability metric. Finally, we rank all feature extractors in descending order of the average. 

To simplify the notation, we denote the aggregated domain's label and feature as $\rvy = (y_1, \ldots, y_n)^\top \in \R^{n}$ and $\Phi = \big( \phi(x_1), \ldots, \phi(x_n) \big)^\top \in \R^{n \times d}$, respectively.
We use $\rvy' \in \R^{n'}$ and $\Phi' \in \R^{n' \times d}$ for the held-out domain.
The main idea of the designed metric is to evaluate whether the classifier fitted on $(\rvy, \Phi)$ also performs well on $(\rvy', \Phi')$.
Hence, we formulate the problem as estimating the likelihood function of $(\rvy', \Phi')$ given $(\rvy, \Phi)$: 
\benrr
p(\rvy', \Phi' \big| \rvy, \Phi ) 
= p(\rvy'| \Phi', \rvy, \Phi ) p(\Phi'| \Phi),
\eenrr
where $p(\rvy'| \Phi', \rvy, \Phi )$ measures \emph{inter-class discriminability} between features $\Phi'$ and labels $\rvy'$, given the aggregated training data. Meanwhile, $p(\Phi'|\Phi)$ measures covariate shift between features $\Phi$ and $\Phi'$, which quantify the \emph{inter-domain stability}. 

Given a hypothetical space $\cF$ of classifiers, we can write $p(\rvy | \Phi ) = \int_{f \in \cF} p(\rvy| \Phi, f) p(f) \mathrm{d}f$.
We consider a linear classifier~\footnote{According to the  Laplace approximation \cite{MacKay1998Laplace},  
if $p(\rvy| \Phi, f)$ is unimodal at $\boldsymbol{\mu}$, we can take Taylor expansion of the log-likelihood at the mode $\log p(\rvy| \Phi, f) \approx \log p(\boldsymbol{\mu}|\Phi, f) - \frac{1}{2} (\rvy-\boldsymbol{\mu})^\top \Lambda (\rvy-\boldsymbol{\mu})$,
where  $\Lambda= - \nabla_{\rvy^\top} \nabla_{\rvy} \log p(\rvy| \Phi, f)\big|_{\rvy=\boldsymbol{\mu}}$.
The quadratic term implies that 
$p(\rvy| \Phi, f)$ can be approximated with a Gaussian distribution.}, i.e. $f \circ \phi (\rvx) = \rvw^\top \phi(\rvx)$ with a Gaussian prior of $\rvw$:
\begin{equation}\label{linearmodel1}
\rvw \sim \cN({\bf 0}, \alpha^{-1} \bbI_d),\quad  \rvy \big|\Phi, \rvw \sim \cN(\Phi\rvw, \beta^{-1} \bbI_n),
\end{equation}
where $\alpha$ and $\beta$ are two positive parameters. 
Figure~\ref{fig:model1} summarizes the model assumptions in \eqref{linearmodel1} with a directed graphical model.
We estimate $\hat \alpha$ and $\hat \beta$ by maximizing the model evidence 
\benrr
p(\rvy|\Phi;\alpha,\beta )=\int_{\rvw\in\R^d} p(\rvy| \Phi, \rvw; \beta) p(\rvw; \alpha) \mathrm{d}\rvw
\eenrr
according to Algorithm~3 in \citet{you2021ranking}
and compute the likelihood of $\rvy'$ as follows: 
\benrr
p(\rvy'| \Phi', \rvy, \Phi; \hat \alpha, \hat \beta) = \frac{p(\rvy', \rvy| \Phi', \Phi; \hat \alpha, \hat \beta )}{p(\rvy|\Phi; \hat \alpha, \hat \beta )}.
\eenrr 
For measuring covariate shift, we approximate the distribution of $\phi(x)$ with a Gaussian distribution $\cN( \hat \mu_{\phi}, \hat \Sigma_\phi)$, where $\hat \mu_{\phi}$ and $\hat \Sigma_\phi$ are estimated from the training data $\Phi.$ Then we compute the density $p(\Phi'|\Phi) = p(\Phi'|\hat \mu_{\phi}, \hat \Sigma_\phi)$ to quantify the covariate shift.

Finally, we compute the density at the logarithmic scale and this defines the proposed metric
\begin{equation}\label{sec31:eq5}
\log \text{ } p(\rvy' | \Phi', \rvy, \Phi) + \log \text{ }  p(\Phi'| \Phi).
\end{equation}
Please refer to Appendix~\ref{App3:sec3} and ~\ref{App3:sec4} for more details.

One distinctive aspect of our selection process is the cross-domain validation, embodied in the first term of \eqref{sec31:eq5}. 
Across different domains, there are domain-invariant and domain-specific features, where overfitting to the latter can severely harm the OoD generalization.   
By evaluating on held-out domains, we are able to filter out models that fixate on domain-specific features. 
To provide theoretical justification, an explicit analysis in the linear regression setting is conducted, where we show that the model with the optimal metric is the one that selects all domain-invariant features. 
Despite the over-simplification, it does reflect the essence of our approach. Due to page limit, the technical details are presented 
in Appendix \ref{app:cross}.

\subsection{Model Ensemble with Feature Selection}\label{sec32}

The top-ranked PTMs in Section~\ref{sec31} are preferred for solving the OoD generalization task.  
To further aggregate different PTMs, we consider assembling the top-ranked feature extractors and rewrite
$
\Phi = \big[ \Phi^{(1)}, \ldots, \Phi^{(k)}\big],
$
where $\Phi^{(i)}$ is the feature matrix from the $i$-th ranked feature extractor.

As we show in experiments, in most cases, aggregating features from multiple models can outperform any single model. However, simply concatenating features inevitably introduces more noise. As found in \cite{ye2021towards},
non-informative but invariant features from training domains may only bring some noise that is irrelevant to the classification problem, and the accumulation of noise hurts the learnability of the OoD generalization task while increasing the memory and computation cost.
Therefore, we propose a feature selection method under the Bayesian linear model framework in Section~\ref{sec31}.

First, we impose a binary mask $\rvz =(\rz_1, \rz_2, \ldots, \rz_d )^\top$ for the weight vector $\rvw = (\rw_1, \rw_2, \ldots, \rw_d)^\top$, where $\rz_i=1$ indicates that $\rw_i$ is an active weight in the top linear model, i.e., $\rw_i\neq 0$, meaning the corresponding feature is informative, while $\rw_i\approx 0$ if $\rz_i=0,$ indicating a noisy feature that should be screened. Therefore the Bayesian feature selection is formulated by estimating the probability $\pi_i$ of $\rz_{i}$ with $\pi_i:= p(\rz_i = 1)$ and $\vpi=\left\{\pi_1,\pi_2,\ldots,\pi_d \right\}.$ 

To facilitate the utility of the mask, we assume that the weights $\{\rw_i\}$ are independent of each other and each weight $\rw_i$ is drawn from either a slab prior or a spike prior \cite{ishwaran2005spike} with the mean of zero: 
\benrr
p(\rw_i|\rz_i,\alpha_{i,1},\alpha_{i,2})=\left\{\begin{array}{ll}
\cN(0, \alpha_{i,1}^{-1}) & \text { if } \rz_i=1; \\
\cN(0, \alpha_{i,2}^{-1}) & \text { if } \rz_i=0.
\end{array}\right.
\eenrr

We make the Bayesian treatment to the linear model in Section~\ref{sec31} by introducing gamma priors for all inverse variance terms:
\benrr
\alpha_{i, 1} \sim \text{Gamma}(\nu_{i,1}, \nu_{i,2}), \quad
\alpha_{i, 2} \sim \text{Gamma}(\nu_{i,3}, \nu_{i,4}), \quad 
\beta \sim \text{Gamma}(\nu_{0,1}, \nu_{0,2}),
\eenrr
and denote all hyper-parameters as $\vnu=\{\nu_{i,j}\}.$ 
In addition, we denote all latent variables as $\boldsymbol{\xi}=\left\{\beta, \{\rw_i,\rz_i, \alpha_{i,1}, \alpha_{i,2}\}_{i=1}^d \right\}$.
Under certain conditions, maximizing marginal likelihood provably leads to consistent selection and obeys Occam’s razor phenomenon~\cite{ghosal2008nonparametric,yang2017bayesian}, and thus screens non-informative features. To estimate $\pi_i$, the maximum marginal likelihood
estimator of $(\vpi, \vnu)$ is given by
\begin{equation}\label{sec33:eq6}
    \hat \vpi, \hat \vnu \,=\, \argmax_{\vpi,\vnu}\ \log \text{ } p(\rvy| \Phi; \vpi,\vnu) \,=\, \argmax_{\vpi, \vnu}\ \log \int_{\vxi} p(\rvy,\vxi | \Phi; \vpi,\vnu) \mathrm{d} \vxi.
\end{equation}
However, the direct maximization of 
\eqref{sec33:eq6} 
is intractable due to the integration over $\vxi$. 
One possible solution is to use EM algorithm \cite{rovckova2014emvs}.
In the E-step, we compute the conditional expectation:
\benrr
\E_{\vxi} \big[ \log \text{ } p(\rvy,\vxi|\Phi; \vpi,\vnu) \big| \rvy, \Phi; \vpi^{old},\vnu^{old} \big].
\eenrr
Notice that evaluating the expectation involves the posterior distribution of $\vxi.$ However in our case, it is not straightforward to obtain an analytical form of the true posterior distribution. We instead approximate it using variational inference \cite{blei2017variational} by introducing a tractable distribution $Q.$
Considering the following objective function:
\benrr
\cL(Q) = \int_{\vxi} Q(\vxi;\vpi,\vnu) \log \frac{p(\rvy,\vxi|\Phi; \vpi,\vnu)}{ Q(\vxi;\vpi,\vnu) } d \vxi,
\eenrr
which is a lower bound of $\log \text{ }  p(\rvy | \Phi; \vpi,\vnu).$
It has been shown the maximizer of $\mathcal{L}(Q)$ is the optimal approximator of $p(\vxi|\rvy, \Phi;\vpi,\vnu)$ under the KL divergence.
To obtain an explicit solution, we consider the classical mean-field family~\cite{blei2017variational}, where variational distribution $Q$ can be factorized into:
\begin{equation}\label{sec33:eq7}
 Q(\vxi)=Q(\beta) \prod^{d}_{i=1}\Big[Q(\rz_{i})Q(\rw_i)Q(\alpha_{i, 1})Q(\alpha_{i, 2})\Big].
\end{equation}
After all variational parameters in 
\eqref{sec33:eq7} 
are updated by running one-step coordinate gradient descent \cite{blei2017variational}, in the M-step, we update $\vpi^{new}$ and $\vnu^{new}$ by maximizing: 
\begin{equation}
\label{sec33:eq8}
\E_{\vxi \sim Q(\vxi;\vpi^{old},\vnu^{old})} \big[ \log \text{ } p(\rvy, \vxi|\Phi; \vpi,\vnu) \big].
\end{equation}
By repeating the E- and M-step, the estimator $(\vpi^{new},\ \vnu^{new})$ converges to an optimal solution. We then screen those variables with converged prior $\pi_{i}$ smaller than the predefined threshold $\tau$. 
The pseudocode is provided in \textbf{Algorithm}~\ref{Pseudocode:feature_selection} to illustrate the main idea of the proposed method, where $\vxi_{k}$ denotes the $k$-th variable in the set $\boldsymbol{\xi}$ and $\vxi_{-k}$ is the subset of all other variables except $\vxi_{k}$. In the E-step, the optimal approximator $Q(\boldsymbol{\xi})$ under the mean-field family takes the tractable form of the expectation of the joint distribution and the optimization of \eqref{sec33:eq8} in M-step is equivalent to substituting with corresponding variational parameters of $Q(\boldsymbol{\xi})$ from E-step. 
Our derivations for variational approximations and prior hyper-parameters optimization are listed in
Appendix~\ref{FS:sec3}.

\begin{algorithm}[t]
    \caption{Pseudocode of Variational EM Algorithm for Bayesian Feature Selection }
    \label{Pseudocode:feature_selection}
    \begin{algorithmic}
        \STATE \textbf{Step 1}: Initialization: $ \prod^{d}_{i=1}Q^{0}(\rw_i)$ and $\prod^{d}_{i=1}Q^{0}(\alpha_{i, 1})Q^{0}(\alpha_{i, 2})$; 
        
        \STATE \textbf{Step 2 (E-Step)}: Approximate the posterior of $\vxi_{k}$ for each $\vxi_{k} \in \boldsymbol{\xi}$ at iteration $t$ with:
        \begin{equation*}
         Q^{t}\left(\vxi_{k}\right)=\exp\left[\E_{Q^{t-1}(\vxi_{-k})}\log \text{ } p(\rvy,\vxi|\Phi;\vpi^{t-1},\vnu^{t-1})\right];
        \end{equation*}
        
        \STATE \textbf{Step 3 (M-Step)}: Update $(\vpi, \vnu)$ at iteration $t$ by maximizing \eqref{sec33:eq8}:
        \begin{equation*}
       {\vpi}^{t}, {\vnu}^{t} = \argmax_{\vpi,\vnu} \text{ } \E_{\vxi \sim Q^{t}(\vxi)} \big[ \log \text{ } p(\rvy, \vxi|\Phi; \vpi,\vnu) \big];
        \end{equation*}
        
       \STATE \textbf{Step 4}: Repeat Step 2 and Step 3 until the convergence criterion is met.
    
    \end{algorithmic}
\end{algorithm}

However, the proposed algorithm still suffers from heavy computational cost: each iteration costs $\mathcal{O}(nd^{2}).$ To address this problem, we propose an efficient version based on stochastic variational inference~\cite{hoffman2013stochastic}. A local estimator $Q^{s}(\vxi)$ is established under stochastic approximation that enjoys less computational complexity and guarantees convergence to global optimum~\cite{robbins1951stochastic}.
We successfully reduce the computation cost to $\mathcal{O}(n^{s}d^{2})$ with $n^{s} \ll n$. Readers can refer to Appendix~\ref{FS:sec4} for more detailed discussions and the complete algorithm for feature selection.

\section{Experiments}\label{experiments}
In this section, we demonstrate the effectiveness of ZooD. First, we evaluate the ability of our ranking metric to estimate OoD performance and compare it with the ground-truth performance and several existing IID ranking methods. Second, we show that our aggregation method achieves significant improvements and SOTA results on several OoD datasets. Finally, we demonstrate that ZooD requires significantly less computation, and, therefore, is practically scalable compared with naive fine-tuning.

\begin{figure*}[t]
    \centering
    \includegraphics[width=0.85\textwidth]{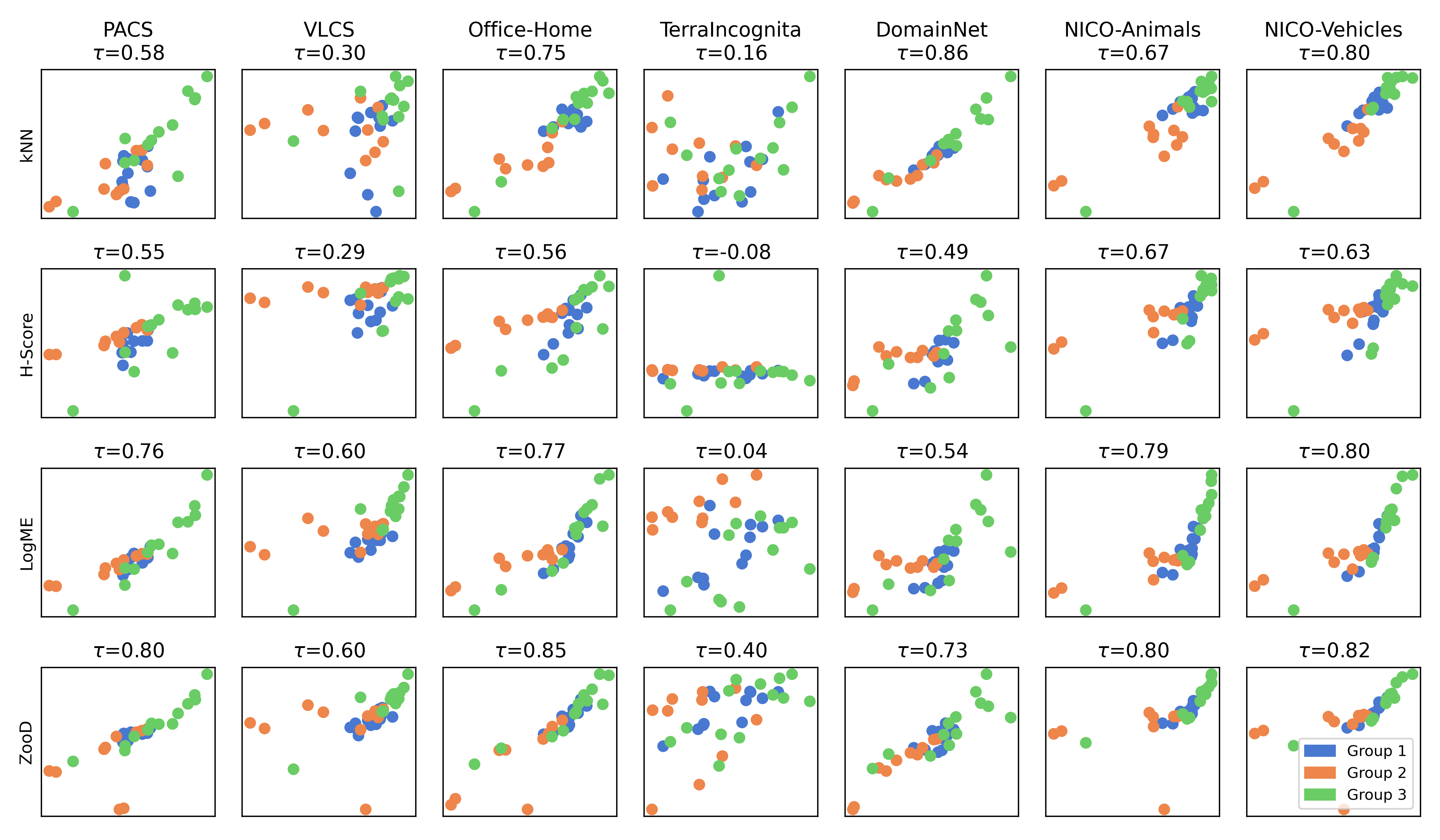}
    \caption{Comparison of ZooD ranking scores with three features-based ranking methods. The plots illustrate ground-truth out-of-domain accuracies ($x$-axis), ranking scores ($y$-axis), and Kendall's coefficient $\tau$ for 35 PTMs on seven datasets. 
    }
    \label{fig:feature_comparison}
    \vspace{-18pt}
\end{figure*}

\textbf{Setup Details.}
We use 35 PTMs with diverse architectures, pre-training methods, and pre-training datasets. 
We divide the PTMs into three groups. Group 1 consists of models with different architectures, Group 2 consists of models pre-trained with different training methods, and Group 3 consists of models pre-trained on large-scale datasets. We conduct experiments on six OoD datasets: PACS~\cite{Li2017DeeperBA}, VLCS~\cite{Fang2013UnbiasedML}, Office-Home~\cite{Venkateswara2017DeepHN}, TerraIncognita~\cite{Beery2018RecognitionIT}, DomainNet~\cite{Peng2019MomentMF}, and NICO (NICO-Animals \& NICO-Vehicles)~\cite{he2020towards}. 
Each of the datasets has multiple domains. The standard way to conduct the experiment is to choose one domain as test (unseen) domain and use the remaining domains as training domains, which is named leave-one-domain-out protocol. The top linear classifier is trained on the training domains only and tested on the test domain. Each domain rotates as the test domain and the average accuracy is reported for each dataset. 
To get ground-truth performance, we follow DomainBed~\cite{gulrajani2020search} to fine-tune top linear classifiers for the PTMs on these OoD datasets. We adopt the leave-one-domain-out cross-validation setup in DomainBed with 10 experiments for hyper-parameter selection and run 3 trials. We triple the number of iterations for DomainNet (5000 to 15000) as it is a large-scale dataset requiring more iterations~\cite{Cha2021SWADDG} and decrease the number of experiments for hyper-parameter selection from 10 to 5. More details on the experimental setup are in Appendix~\ref{sec:supp_setup}.

\subsection{Comparison with IID Ranking Metrics} \label{sec:compare_iid}

\textbf{IID ranking methods.}
We divide existing ranking methods into two groups. One group consists of methods that employ PTM's classification layer for ranking. These methods include NCE~\cite{Tran2019TransferabilityAH} and LEEP~\cite{Nguyen2020LEEPAN}. The other group consists of approaches that only use PTM's extracted features. These methods include H-Score~\cite{Bao2019AnIA} and LogME~\cite{you2021logme}. Additionally, we also use kNN with k=200~\cite{Wu2018UnsupervisedFL} as a baseline.

\textbf{Evaluation metrics.}
To evaluate PTMs on OoD datasets with ranking methods, we follow leave-one-domain-out validation protocol~\cite{Li2017DeeperBA}. For ZooD and kNN, we further adopt leave-one-domain-out validation for training domains and take average results as the performance prediction for the held-out test domain. To compute the correlation between ranking scores and ground-truth performance, we use two metrics. First, to compare the ranking of a transferability metric with accuracy, we employ Kendall's coefficient $\tau$~\cite{kendall1938new}. Unlike Pearson's correlation, $\tau$ measures correlation based on the order of two measures. Consequently, it is a better criterion for ranking.
Second, to measure the performance of transferability metric for top-model selection, we utilize weighted Kendall's coefficient $\tau_w$~\cite{vigna2015weighted}. The $\tau_w$ gives more weight to the ranking of top-performing models compared with the rest of the models. Therefore, it is a better comparative criterion for top model selection.

\begin{figure*}[t]
    \centering
    \includegraphics[width=0.9\textwidth]{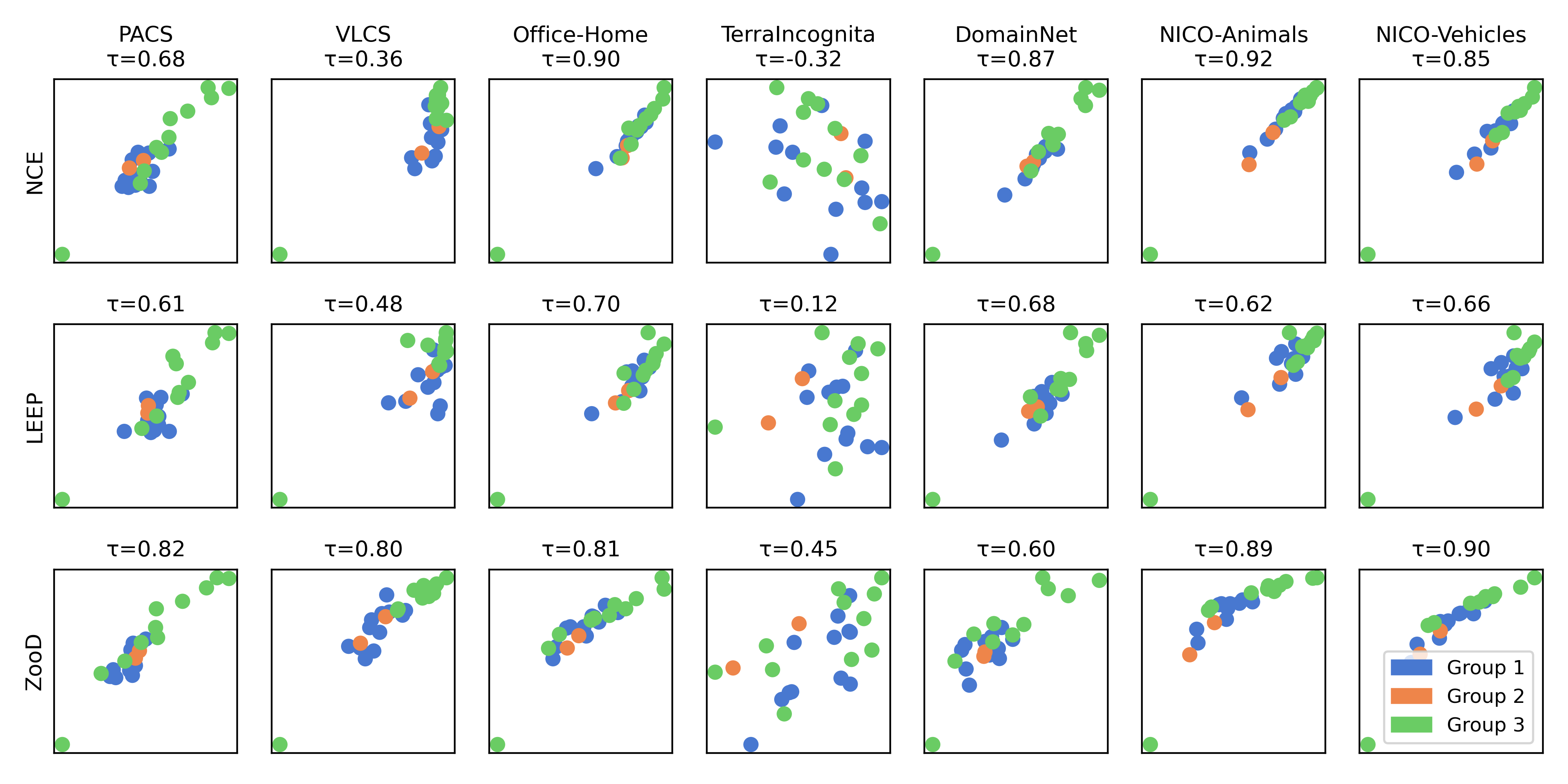}
    \caption{Comparison of ZooD ranking scores with two classification-layer-based ranking methods. The plots illustrate ground-truth out-of-domain accuracies ($x$-axis), ranking scores ($y$-axis), and Kendall's coefficient $\tau$ for 25 PTMs that have classification layers on seven datasets.}
    \label{fig:compare_classification}
    \vspace{-10pt}
\end{figure*}

\begin{table*}[t]
\caption{ Comparisons: (a) $\tau_w$ between ZooD and feature-based transferability estimation methods using all of our PTMs. (b) $\tau_w$ between ZooD and classification-based transferability estimation methods. For this comparison, we consider 25 models that have classification heads. (c) Our method v.s. brute-force fine-tuning in terms of computing cost. For this comparison, we consider all 35 models.}
\label{all_comparisons}
\begin{subtable}[t]{0.35\textwidth}
\caption{$\tau_w$ for feature based}
\label{tab:feature_based_mini}
\resizebox{0.99\textwidth}{!}{
\begin{tabular}{lrrrr}
\toprule
{} &  kNN &  H-Score &  LogME &  ZooD \\
\midrule
PACS           & 0.76 &     0.57 &   0.88 &  \textbf{0.91} \\
VLCS           & 0.49 &     0.45 &   0.79 &  \textbf{0.80} \\
Office-Home    & 0.78 &     0.68 &   \textbf{0.86} &  \textbf{0.86} \\
TerraIncognita & 0.40 &    -0.20 &   0.02 &  \textbf{0.46} \\
DomainNet      & \textbf{0.89} &     0.62 &   0.65 &  0.76 \\
NICO-Animals   & 0.73 &     0.72 &   0.89 &  \textbf{0.90} \\
NICO-Vehicles  & 0.82 &     0.75 &   0.90 &  \textbf{0.92} \\
\bottomrule
\end{tabular}}
\end{subtable}
\hfill
\begin{subtable}[t]{0.27\textwidth}
\caption{$\tau_w$ for Classification based}
\label{tab:classification_based_mini}
\resizebox{0.99\textwidth}{!}{
\begin{tabular}{lrrr}
\toprule
{} &  LEEP &   NCE &  ZooD \\
\midrule
PACS           &  0.76 &  0.81 &  \textbf{0.89} \\
VLCS           &  0.57 &  0.32 &  \textbf{0.88} \\
Office-Home    &  0.76 &  \textbf{0.94} &  0.86 \\
TerraIncognita &  0.02 &  -0.44 &  \textbf{0.59} \\
DomainNet      &  0.77 &  \textbf{0.87} &  0.72 \\
NICO-Animals   &  0.58 &  0.92 &  \textbf{0.94} \\
NICO-Vehicles  &  0.69 &  0.92 &  \textbf{0.95} \\
\bottomrule
\end{tabular}}
\end{subtable}
\hfill
\begin{subtable}[t]{0.33\textwidth}
\caption{Speed-up over brute-force}
\label{tab:computing_cost_mini}
\resizebox{0.99\textwidth}{!}{
\begin{tabular}{lcrr}
\toprule
GPU Hours &  ZooD &  Fine-tuning & Speed Up\\
\midrule
PACS           & 0.27 & 2662.27 & 9859$\times$ \\
VLCS           & 0.29 & 2706.67 & 9332$\times$ \\
Office-Home    & 0.39 & 3089.87 & 7922$\times$ \\
TerraIncognita & 0.49 & 3920.27 & 8000$\times$ \\
DomainNet      & 11.24 & 17055.33 & 1516$\times$ \\
NICO-Animals   & 0.32 & 2914.40 & 9107$\times$ \\
NICO-Vehicles  & 0.30 & 2794.13 & 9313$\times$ \\
\bottomrule
\end{tabular}}
\end{subtable}
\vspace{-15pt}
\end{table*}

\textbf{Results.} 
First, we compare our method with feature-based scoring methods: kNN, H-Score, and LogME. These methods, similar to our method, rank models based on the penultimate layer. We compare ZooD with these methods for the full set of 35 PTMs. We plot ranking scores and ground-truth accuracies in Figure~\ref{fig:feature_comparison}. For quantitative comparison, we also provide $\tau$ values. 
It can be seen that ZooD is better correlated with fine-tuning accuracy than other ranking methods on most of the datasets. For example, our method has a $\tau$ of 0.85 compared with LogME's $\tau$ of 0.77 on Office-Home and a $\tau$ of 0.40 compared with LogME's $\tau$ of 0.04 on TerraIncognita.

Furthermore, our metric is more stable and consistent. Precisely, $\tau$ of ZooD varies between 0.40 {\small $\sim$} 0.85 compared with 0.04 {\small $\sim$} 0.80 for LogME, -0.08 {\small $\sim$} 0.67 for H-Score, and 0.16 {\small $\sim$} 0.86 for kNN. The consistency of transferability metric across different datasets is critical since the purpose of a transferability metric is to estimate performance on a new dataset without having access to ground-truth accuracy. Whenever an estimation metric is inherently unstable, it is hard to determine its reliability for a new dataset.

Note that our method uses a linear model with Gaussian error to approximate the top classifier. This helps us achieve efficient model assessment, especially on small and medium-sized datasets in which the bias caused by model approximation is negligible compared with the estimation error due to insufficient data. However, on DomainNet, things may be different. The bias caused by model approximation dominants the evaluation performance on large datasets. Therefore, our method does not outperform kNN on DomainNet.

Second, we compare our method with classification-layer based methods: NCE and LEEP. For this comparison, we select a subset of our PTMs that have classification layers. The results are illustrated in Figure~\ref{fig:compare_classification}. It can be seen that ZooD is also more stable and consistent than NCE and LEEP. 
Moreover, Our method achieves superior performance on the difficult real-world TerraIncognita dataset. This dataset consists of obscure and blurry images captured by WildCams installed in different territories. NCE has a negative correlation for this dataset. On the other hand, our method, although not perfect, captures the relation in a better way. For this challenging dataset, our method has a $\tau$ of 0.45 compared with 0.12 and -0.32 for LEEP and NCE, respectively.

Third, we compare the weighted Kendall's coefficient of our method with other ranking methods. The weighted Kendall's coefficient is a better metric to gauge the performance of a metric for top model selection. 
We also divide these results into two groups: comparison with feature-based scoring methods in Table~\ref{tab:feature_based_mini} and comparison with classification-based scoring methods in Table~\ref{tab:classification_based_mini}. 
Our method outperforms feature-based scoring methods on 6 out of 7 datasets. Similarly, it also outperforms both LEEP and NCE on 5 out of 7 datasets.
Moreover, our ranking method is more stable as it performs better on challenging datasets. For example, it has $\tau_w$ of 0.46 {\small $\sim$} 0.92 compared with LogME's $\tau_w$ of 0.02 {\small $\sim$} 0.90 and H-Score's $\tau_w$ of -0.20 {\small $\sim$} 0.75.

In summary, transferability estimation of ZooD correlates better with ground-truth accuracy on most of the OoD datasets compared with previous ranking methods. It also outperforms most feature-based metrics for model selection in terms of $\tau_w$. Additionally, it is more stable and consistent across datasets, making it a better choice for pre-trained model selection. 

\begin{table}
    \centering
    \caption{Comparison of out-of-domain accuracies between ZooD and SOTA OoD methods. The results of MixStyle~\cite{Zhou2021DomainGW} and SWAD~\cite{Cha2021SWADDG} are from SWAD, and other results are from~\citet{gulrajani2020search} (denoted with ${\dagger}$). Our results are average of three trials.}
    \label{tab:sota_results}
    \resizebox{0.85\textwidth}{!}{
    \begin{tabular}{l|ccccc|c}
        \toprule
        \textbf{Method} &  \textbf{PACS}  & \textbf{VLCS}  & \textbf{Office-Home} & \textbf{TerraInc.} & \textbf{DomainNet}   & \textbf{Avg}  \\
        \hline
        ERM$^{\dagger}$           & 85.5 & 77.5 & 66.5 & 46.1 & 40.9 & 63.3 \\ 
        IRM$^{\dagger}$           & 83.5 & 78.6 & 64.3 & 47.6 & 33.9 & 61.6 \\ 
        GroupDRO$^{\dagger}$      & 84.4 & 76.7 & 66.0 & 43.2 & 33.3 & 60.7 \\ 
        I-Mixup$^{\dagger}$       & 84.6 & 77.4 & 68.1 & 47.9 & 39.2 & 63.4 \\ 
        MLDG$^{\dagger}$          & 84.9 & 77.2 & 66.8 & 47.8 & 41.2 & 63.6 \\ 
        MMD$^{\dagger}$           & 84.7 & 77.5 & 66.4 & 42.2 & 23.4 & 58.8 \\ 
        DANN$^{\dagger}$   & 83.7 & 78.6 & 65.9 & 46.7 & 38.3 & 62.6 \\ 
        CDANN$^{\dagger}$                                     & 82.6 & 77.5 & 65.7 & 45.8 & 38.3 & 62.0 \\ 
        MTL$^{\dagger}$           & 84.6 & 77.2 & 66.4 & 45.6 & 40.6 & 62.9 \\ 
        SagNet$^{\dagger}$        & 86.3 & 77.8 & 68.1 & 48.6 & 40.3 & 64.2 \\ 
        ARM$^{\dagger}$           & 85.1 & 77.6 & 64.8 & 45.5 & 35.5 & 61.7 \\ 
        VREx$^{\dagger}$     & 84.9 & 78.3 & 66.4 & 46.4 & 33.6 & 61.9\\ 
        RSC$^{\dagger}$           & 85.2 & 77.1 & 65.5 & 46.6 & 38.9 & 62.7 \\ 
        MixStyle      & 85.2 & 77.9 & 60.4 & 44.0 & 34.0 & 60.3 \\ 
        SWAD           & 88.1 & 79.1 & 70.6 & 50.0 & 46.5 & 66.9 \\
        \hline
        \multicolumn{7}{c}{ZooD} \\
         \hline
         Single   & 96.0 & 79.5 & 84.6 & 37.3 & 48.2 & 69.1 \\
         Ensemble & 95.5 & 80.1 & 85.0 & 38.2 & 50.5 & 69.9 \\
         F. Selection & \textbf{96.3} & \textbf{80.6} & \textbf{85.1} & 42.3 & \textbf{50.6} & \textbf{71.0} \\
         \hline
         F. Ratio (\% ) & 24.3 & 24.5 & 62.5 & 76.8 & 99.8 \\
         \bottomrule
    \end{tabular}
    }
\vspace{-10pt}
\end{table}

\subsection{SOTA Results with Our Selection Method}\label{sota_results}

We also compare ZooD (model ranking and feature selection) with several recent SOTA OoD methods and demonstrate that it achieves substantial performance improvements. We compare previous OoD methods with three versions of our method: 1) \textbf{Single}: fine-tune the top-1 model by transferability metric; 2) \textbf{Ensemble}:  fine-tune an ensemble of the top-K models; 3) \textbf{F. Selection}: fine-tune an ensemble of the top-K models with feature selection, which is the expected result using ZooD. By fine-tuning, we mean using ERM with DomainBed settings to fine-tune a top linear classifier for the PTMs. Their predictive performance and \textbf{F. Ratio} (the percentage of features used in \textbf{F. Selection}) are listed in the last four lines of Table~\ref{tab:sota_results}.

In all experiment results, except TerraIncognita (discussed in the next paragraph), our method achieves remarkable improvement against ERM and recent SOTA. For \textbf{Single}, we list the improvements over the previous SOTA as follows: +14\% on Office-Home, +7.9\% on PACS, +1.7\% on DomainNet, and +0.4\% on VLCS. 
This result also shows that even without aggregation, using proper pre-trained model can improve OoD generalization by a large margin. Notice that our method can be regarded as a complement to other OoD algorithms. After selecting the top-ranked models, different OoD algorithms can be adapted to fine-tune the models.

The performance of \textbf{Single} does not outperform the previous SOTA on TerraIncognita. This is because previous methods fine-tune the whole network. In contrast, we only train a classifier on top of a fixed feature extractor. TerraIncognita is a much more challenging dataset compared with other OoD datasets, as the majority of its images are obscured by the background. Therefore it requires fully fine-tuning. To show the effectiveness of ZooD with fully fine-tuning, we select the top-1 ranked model and fine-tune the whole model. Our resulted model achieves a +2.6\% improvement compared with the previous SOTA. One limitation of ZooD when aggregating multiple models is that fine-tuning the whole models is difficult due to the limitation of GPU memory. However, for OoD tasks, fine-tuning the whole model may not perform better than fine-tuning the top classifier. For example, the results of fine-tuning the full top-ranked models on PACS, VLCS and Office-Home are 90.6, 79.1 and 83.4, respectively. Empirically, we find if a PTM is suitable for a given OoD task, fine-tuning the top classifier has better OoD generalization than fine-tuning the full model.

As shown in Table~\ref{tab:sota_results}, a simple model ensemble (\textbf{Ensemble}) provides fairly minimal improvement because it may introduce invariant but noisy features. To efficiently utilize multiple models, we propose to select informative features in Section~\ref{sec32}. Here, we compare the performance improvement by \textbf{F. Selection} with \textbf{Single} and \textbf{Ensemble}. ZooD significantly outperforms both candidates while only using a small portion of aggregated features from top-K models. Even on the most sophisticated DomainNet, 
ZooD can improve predictive performance by +2.4\% compared with \textbf{Single} and +0.1\% compared with \textbf{Ensemble}.

\begin{wrapfigure}{r}{0.48\textwidth}
    \centering
    \includegraphics[width=0.48\textwidth]{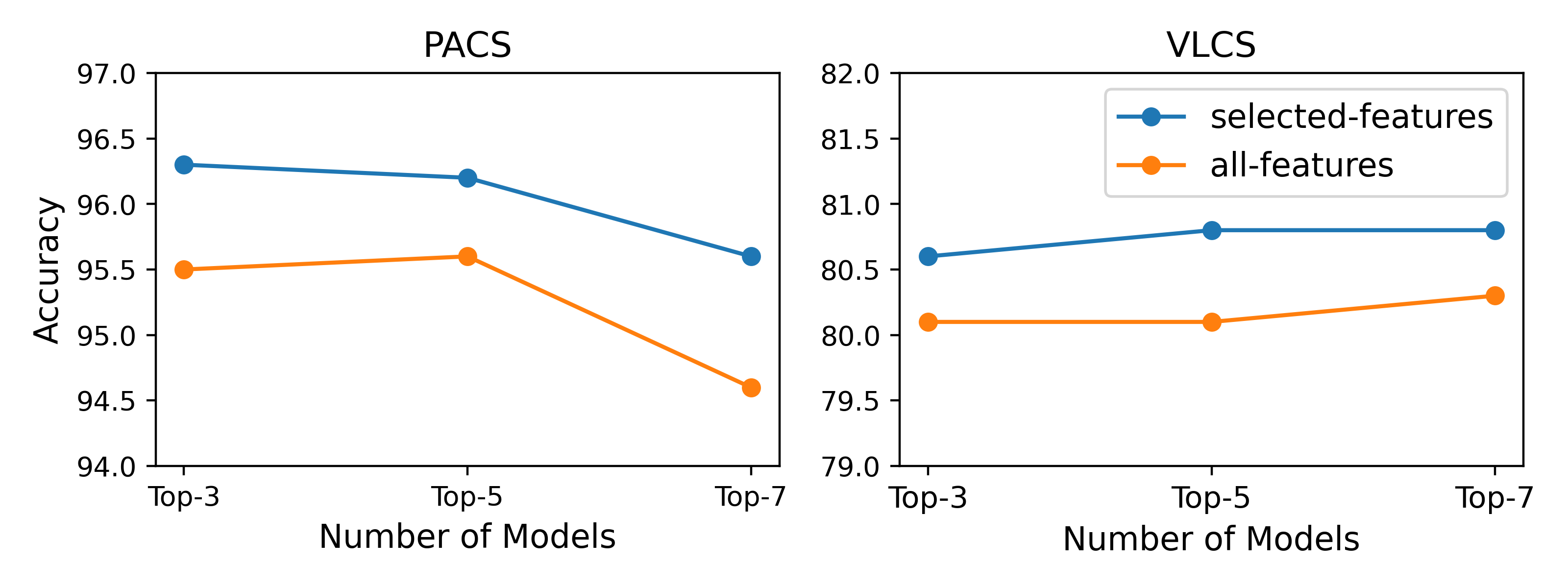}
    \caption{Comparison of selected-feature ensemble vs. all-feature ensemble for varying number of top models in the ensemble.}
    \label{fig:abalation1}
    \vspace{-10pt}
\end{wrapfigure}

To find the appropriate number K for the model ensemble, we performed an ablation study. We varied the number of K, e.g. $\text{K}\in\{3, 5, 7\}$. The performance changes are plotted in Figure~\ref{fig:abalation1}. We found the performance by aggregating top-3 models strikes the right balance between performance and computational complexity. Hence, $\text{K}=3$ is set to the default value.

In summary, our ranking metric in ZooD is good enough to select a model that can outperform the previous SOTA methods without adding any bells and whistles. Furthermore, feature selection in ZooD can efficiently utilize informative features from top-K models to further improve OoD generalization. In this work, we do not control for the impact of better PTMs. Given a zoo of PTMs, our method aims to exploit the power of the zoo for OoD generalization. We can further increase the power of the model zoo by adding more PTMs. Based on extensive experimental results on various OoD datasets, we conclude ZooD makes it easy and efficient to exploit a large set of PTMs for OoD generalization.

\subsection{Computational Efficiency of ZooD}
We illustrate the precision and computational efficiency of ZooD by comparing it with brute-force fine-tuning in terms of GPU hours.
The results are shown in Table~\ref{tab:computing_cost_mini}. ZooD provides a minimum of 1516$\times$ speed-up for DomainNet and a maximum of 9859$\times$ speed-up for PACS. Cumulatively, our method took a total of 13 GPU hours to evaluate all the PTMs on all the datasets compared with 35140 GPU hours (equivalent to 4 GPU years) for brute-force fine-tuning. Therefore, ZooD is a scalable and 
practical method for OoD generalization.

\section{Conclusion}
Machine learning models rely on IID assumption, which is often violated due to constant distribution shifts in real-world applications. In this work, we argue for leveraging a large set of PTMs to improve OoD generalization and propose ZooD, a paradigm for efficient PTMs ranking and aggregation. 
Our paradigm avoids the computationally-prohibitive fine-tuning by ranking PTMs based on quantifying their inter-class discriminability and inter-domain stability, and selecting the most informative features from top-ranked PTMs ensemble. Extensive experiments show ZooD is superior in ranking correlation with the ground-truth performance and achieves SOTA results on various OoD benchmarks.

\vspace{-10pt}
\section*{Acknowledgments}
\vspace{-10pt}
Awais's work, in part, was also supported by the Institute of Information \& Communications Technology Planning \& Evaluation (IITP) grant (No. 2021-0-02068, Artificial Intelligence Innovation Hub and No.RS-2022-00155911, Artificial Intelligence Convergence Innovation Human Resources Development (Kyung Hee University)).
We gratefully acknowledge the support of MindSpore for this research.

{\small
\bibliographystyle{plainnat}

\bibliography{refs}
}

\newpage
\appendix

\part*{Appendix}

\input{appendix}

\end{document}

%% file: xcl_math.tex
\usepackage{amsmath,amssymb}
\usepackage{graphicx}
\usepackage{color}
\usepackage{bm}
\usepackage{algorithm}
\usepackage{algorithmic}
\usepackage{wrapfig}
\usepackage{hyperref}
\usepackage{url}

\newtheorem{theorem}{Theorem}

\newcommand{\benr}{\begin{eqnarray}}
\newcommand{\eenr}{\end{eqnarray}}
\newcommand{\benrr}{\begin{eqnarray*}}
\newcommand{\eenrr}{\end{eqnarray*}}
\newcommand{\ben}{\begin{equation}}
\newcommand{\een}{\end{equation}}
\newcommand{\benn}{\begin{equation*}}
\newcommand{\eenn}{\end{equation*}}

\newcommand{\cD}{\mathcal D}

\newcommand{\cF}{\mathcal F}

\newcommand{\cL}{\mathcal L}
\newcommand{\cN}{\mathcal N}
\newcommand{\cM}{\mathcal M}

\newcommand{\cX}{\mathcal X}

\newcommand{\bbI}{\mathbb I}

\def\rw{{\textnormal{w}}}

\def\rz{{\textnormal{z}}}

\def\rvm{{\mathbf{m}}}

\def\rvv{{\mathbf{v}}}
\def\rvw{{\mathbf{w}}}
\def\rvx{{\mathbf{x}}}
\def\rvy{{\mathbf{y}}}
\def\rvz{{\mathbf{z}}}

\newcommand{\E}{\mathbb{E}}

\newcommand{\R}{\mathbb R}

\DeclareMathOperator*{\argmax}{arg\,max}

\def\rmA{{\mathbf{A}}}

%% file: appendix.tex
\section{Experiments}\label{sec:supp_experiments}

\subsection{Complete Details of Experiment Setup}\label{sec:supp_setup}
In this section, we provide a detailed experiment setup we have used. For completeness purposes, this section also includes details already mentioned in the main paper. 

\textbf{Pre-trained models.} We use 35 PTMs having diverse architectures, pre-training methods and pre-training datasets.
\textbf{Group 1} consists of models with different architectures. This group consists of 12 different architectures (CNNs and ViTs) trained on ImageNet-1k. The architectures are as follows: ResNet-50, ResNet-152~\cite{He2016DeepRL}, 
ResNeXt-50~\cite{Xie2017AggregatedRT}, DenseNet-169, DenseNet-201~\cite{Huang2017DenselyCC}, Inception v1~\cite{Szegedy2015GoingDW}, Inception v3~\cite{Szegedy2016RethinkingTI}, MobileNet v2~\cite{Sandler2018MobileNetV2IR}, EfficientNet-B2, EfficientNet-B4~\cite{Tan2019EfficientNetRM}, Swin-T, Swin-B~\cite{Liu2021SwinTH}.
\textbf{Group 2} consists of models pre-trained with different training methods. We use 10 ResNet-50s trained via following pre-training methods: Adversarial Training~\cite{Madry2018TowardsDL}, BYOL~\cite{Grill2020BootstrapYO}, MoCo-v2~\cite{Chen2020ImprovedBW}, InsDis~\cite{Wu2018UnsupervisedFL}, PIRL~\cite{Misra2020SelfSupervisedLO}, DeepCluster-v2~\cite{Caron2018DeepCF}, PCL-v2~\cite{Li2021PrototypicalCL}, SeLa-v2~\cite{Asano2020SelflabellingVS, Caron2020UnsupervisedLO}, SwAV~\cite{Caron2020UnsupervisedLO}.
\textbf{Group 3} consists of models pre-trained on large-scale datasets. We used 13 different models trained on ImageNet-22k~\cite{Russakovsky2015ImageNetLS}, YFCC-100M~\cite{Thomee2016YFCC100MTN}, IG-1B-Targeted~\cite{Yalniz2019BillionscaleSL}, WebImageText~\cite{Radford2021LearningTV}.
A summary of the PTMs can be found in Table~\ref{tab_supp:pre-trained_models}.

\textbf{Datasets.} We use six OoD datasets for our experiments. The details of these datasets are listed here. PACS~\cite{Li2017DeeperBA} consists of 9,991 images from four domains (art, cartoons, photos, sketches) and seven classes. VLCS~\cite{Fang2013UnbiasedML} consists of 10,729 images from four domains (Caltech101, LabelMe, SUN09, VOC2007) and five classes. Office-Home~\cite{Venkateswara2017DeepHN} has four domains (art, clipart, product, real) of common objects in office and home settings. The dataset has a total of 15,588 images belonging to 65 classes. TerraIncognita~\cite{Beery2018RecognitionIT} contains photos of wild animals taken by camera traps installed at four different locations. It has a total of 24,788 images from 10 classes. DomainNet~\cite{Peng2019MomentMF} is one of the most challenging OoD datasets. It has 586,575 images from six diverse domains (clipart, infographics, painting, quickdraw, real, sketch) belonging to 345 classes. NICO~\cite{he2020towards} consists of nearly 25,000 images from two superclasses: NICO-Animals (10 classes) and NICO-Vehicles (9 classes). We split the images of NICO-Animals and NICO-Vehicles into multiple domains according to~\cite{bai2020decaug} and combine validation and test sets as one domain to form four domains, separately.

\textbf{Ground-truth performance.} To get ground-truth performance, we train linear classifiers on top of PTMs following DomainBed~\cite{gulrajani2020search}. The authors of DomainBed~\cite{gulrajani2020search} argue for the hyper-parameter selection to be a part of the method selection criteria. Based on this argument, they propose a rigorous test bench. We follow their training and evaluation protocol, including dataset splits, hyper-parameter settings, optimizer, etc. We adopt the leave-one-domain-out cross-validation setup in DomainBed with 10 experiments for hyper-parameter selection and run 3 trials. We triple the number of iterations for DomainNet (5000 to 15000) as it is a larger dataset and requires more training~\cite{Cha2021SWADDG} and decrease the number of experiments for hyper-parameter selection from 10 to 5.

\textbf{IID ranking methods.}
We divide existing ranking methods into two groups. The first group consists of methods that employ PTM's classification layer for ranking. These methods include NCE~\cite{Tran2019TransferabilityAH} and LEEP~\cite{Nguyen2020LEEPAN}. The second group consists of approaches that only use PTM's extracted features. These methods include H-Score~\cite{Bao2019AnIA} and LogME~\cite{you2021logme}. Additionally, we also use kNN with k=200~\cite{Wu2018UnsupervisedFL} as a baseline.

\textbf{Evaluation metrics.}
To evaluate PTMs on OoD datasets with ranking methods, we follow leave-one-domain-out validation protocol~\cite{Li2017DeeperBA}. For ZooD and kNN, we further adopt leave-one-domain-out validation for training domains and take average results as the performance prediction for the held-out test domain. To compute the correlation between ranking scores and ground-truth performance, we use two metrics. First, to compare the ranking of a transferability metric with accuracy, we employ Kendall's coefficient $\tau$~\cite{kendall1938new}. Unlike Pearson's correlation, $\tau$ measures correlation based on the order of two measures. Consequently, it is a better criterion for ranking.
Second, to measure the performance of transferability metric for top-model selection, we utilize weighted Kendall's coefficient $\tau_w$~\cite{vigna2015weighted}. The $\tau_w$ gives more weight to the ranking of top-performing models compared with the rest of the models. Therefore, it is a better comparative criterion for top model selection.

\subsection{Extended Ranking Results}

In this section, we provide detailed and raw results for all 35 models on all six OoD datasets. Specifically, we provide raw scores assigned by all the ranking methods to all PTMs. We also provide accuracy of each model after fine-tuning. A more interpretable and visual analysis of these scores are provided in section~\ref{sec:compare_iid} of the main paper. 

We provide these raw scores here to help aid reproducability and to help other researchers for easier benchmarking. It is important to note that getting these results, especially accuracy results, is computationally expensive, which may hinder future progress. 
For instance, on large DomainNet dataset, it takes 711 GPU days of training to get all ground-truth performance. Therefore, providing these raw scores can significantly help future researchers. 

The results are provided in the following tables. Table~\ref{tab:full_result_pacs_vlcs} shows results on PACS and VLCS, Table~\ref{tab:full_result_officehome_terraincognita} shows results on Office-Home and TerraIncognita, Table~\ref{tab:full_result_nico} contains results on NICO-Animals and NICO-Vehicles, and Table~\ref{tab:full_result_domainnet} contains results on DomainNet.

\begin{table}[h]
    \centering
    \caption{Details of our model zoo. The first column corresponds to the numbers we have used for subsequent tables. The rest of the table describes architectures, pre-training datasets, and pre-training algorithms as well as the group and source of each model.}
    \label{tab_supp:pre-trained_models}
    \begin{adjustbox}{width=0.95\textwidth}
        \begin{tabular}{clllll}
        \toprule
        Number & Architecture & Dataset & Algorithm & Group & Source  \\
        \hline
        1 & ResNet-50 & ImageNet-1K    & ERM & Group 1 & \citet{NEURIPS2019_9015}\\
        2 & ResNet-152 & ImageNet-1K   & ERM & Group 1 & \citet{NEURIPS2019_9015}\\
        3 & ResNeXt-50 & ImageNet-1K   & ERM & Group 1 & \citet{NEURIPS2019_9015}\\
        4 & DenseNet-169 & ImageNet-1K & ERM & Group 1 & \citet{NEURIPS2019_9015}\\
        5 & DenseNet-201 & ImageNet-1K & ERM & Group 1 & \citet{NEURIPS2019_9015}\\
        6 & Inception v1 & ImageNet-1K & ERM & Group 1 & \citet{NEURIPS2019_9015} \\
        7 & Inception v3 & ImageNet-1K & ERM & Group 1 & \citet{NEURIPS2019_9015}\\
        8 & MobileNet v2 & ImageNet-1K & ERM & Group 1 & \citet{NEURIPS2019_9015}\\
        9 & EfficientNet-B2 & ImageNet-1K & ERM & Group 1 & \citet{NEURIPS2019_9015}\\
        10 & EfficientNet-B4 & ImageNet-1K & ERM & Group 1 & \citet{NEURIPS2019_9015} \\
        11 & Swin-T & ImageNet-1K & Swin & Group 1 & \citet{Liu2021SwinTH}\\
        12 & Swin-B & ImageNet-1K & Swin & Group 1 & \citet{Liu2021SwinTH}\\
        \hline
        13 & ResNet-50 & ImageNet-1K & Adv. $\ell_2$ ($\epsilon=0.5$) & Group 2& \citet{Salman2020DoAR} \\
        14 & ResNet-50 & ImageNet-1K & Adv. $\ell_\infty$ ($\epsilon=4$) & Group 2& \citet{Salman2020DoAR} \\
        15 & ResNet-50 & ImageNet-1K & BYOL & Group 2 & \citet{Ericsson2021HowTransfer}\\
        16 & ResNet-50 & ImageNet-1K & MoCo-v2 & Group 2 & \citet{Ericsson2021HowTransfer}\\
        17 & ResNet-50 & ImageNet-1K & InsDis & Group 2 & \citet{Ericsson2021HowTransfer}\\
        18 & ResNet-50 & ImageNet-1K & PIRL & Group 2 & \citet{Ericsson2021HowTransfer}\\
        19 & ResNet-50 & ImageNet-1K & DeepCluster-v2 & Group 2 & \citet{Ericsson2021HowTransfer}\\
        20 & ResNet-50 & ImageNet-1K & PCL-v2 & Group 2 &  \citet{Ericsson2021HowTransfer}\\
        21 & ResNet-50 & ImageNet-1K & SeLa-v2 & Group 2 & \citet{Ericsson2021HowTransfer}\\
        22 & ResNet-50 & ImageNet-1K & SwAV & Group 2 & \citet{Ericsson2021HowTransfer}\\
        \hline
        23 & ResNet-18   & ImageNet-1K + YFCC-100M  & Semi-supervised & Group 3 & \citet{Yalniz2019BillionscaleSL} \\
        24 & ResNet-50   & ImageNet-1K + YFCC-100M  & Semi-supervised & Group 3 & \citet{Yalniz2019BillionscaleSL} \\
        25 & ResNeXt-50  & ImageNet-1K + YFCC-100M  & Semi-supervised & Group 3 & \citet{Yalniz2019BillionscaleSL} \\
        26 & ResNeXt-101 & ImageNet-1K + YFCC-100M  & Semi-supervised & Group 3 & \citet{Yalniz2019BillionscaleSL} \\
        \hline
        27 & ResNet-18   & ImageNet-1K + IG-1B-Targeted & Semi-weakly Supervised & Group 3    & \citet{Yalniz2019BillionscaleSL}\\
        28 & ResNet-50   & ImageNet-1K + IG-1B-Targeted & Semi-weakly Supervised & Group 3    & \citet{Yalniz2019BillionscaleSL}\\
        29 & ResNeXt-50  & ImageNet-1K + IG-1B-Targeted & Semi-weakly Supervised & Group 3    & \citet{Yalniz2019BillionscaleSL}\\
        30 & ResNeXt-101 & ImageNet-1K + IG-1B-Targeted & Semi-weakly Supervised & Group 3    & \citet{Yalniz2019BillionscaleSL}\\
        \hline
        31 & Swin-B & ImageNet-1K + ImageNet-22K & Swin & Group 3 & \citet{Liu2021SwinTH}\\
        32 & BEiT-B   & ImageNet-1K + ImageNet-22K & BEiT & Group 3 & \citet{wolf-etal-2020-transformers,bao2021beit}\\
        33 & ViT-B/16 & ImageNet-1K + ImageNet-22K & ViT & Group 3 & \citet{wolf-etal-2020-transformers,wu2020visual}\\
        \hline
        34 & ResNet-50 & WebImageText & CLIP & Group 3 & \citet{Radford2021LearningTV}\\
        35 & ViT-B/16 & WebImageText & CLIP & Group 3 & \citet{Radford2021LearningTV}\\
        \bottomrule
        \end{tabular}
    \end{adjustbox}
\end{table}

\newpage

\begin{table}[]
    \centering
    \caption{The ranking scores and fine-tuning accuracy on PACS and VLCS datasets. The numbering in the first column corresponds to a pre-trained model from Table~\ref{tab_supp:pre-trained_models}. The numbers in each subsequent column represent the scores assigned by a ranking metric to the PTMs. The last column displays the accuracy of each model after fine-tuning. Empty cells represent models for which ranking is not feasible.}
    \label{tab:full_result_pacs_vlcs}
    \begin{adjustbox}{width=0.95\textwidth}
        \begin{tabular}{c|ccccccc|ccccccc}
        \toprule
        Model & \multicolumn{7}{c|}{PACS} & \multicolumn{7}{c}{VLCS} \\
        \hline
        Number & LEEP & NCE & H-Score & kNN & LogME & ZooD & Acc. & LEEP & NCE & H-Score & kNN & LogME & ZooD & Acc. \\
        \hline
        1 & -1.226  & -1.077  & 5.016  & 49.608  & 0.226 & 0.053 & 66.9 & -0.566 & -0.498 & 3.241 & 58.156 & 0.223 & 0.119 & 76.7  \\
        2 & -1.140  & -1.007  & 5.072  & 54.767  & 0.274 & 0.100 & 74.4 & -0.538 & -0.494 & 3.253 & 61.215 & 0.229 & 0.127 & 77.0  \\
        3 & -1.185  & -1.022  & 5.010  & 50.737  & 0.231 & 0.064 & 65.6 & -0.552 & -0.499 & 3.216 & 58.540 & 0.200 & 0.083 & 76.9  \\
        4 & -1.156  & -0.998  & 4.636  & 43.284  & 0.186 & -0.012 & 67.1 & -0.569 & -0.514 & 3.013 & 56.056 & 0.181 & 0.063 & 76.8 \\
        5 & -1.172  & -1.039  & 4.854  & 48.861  & 0.235 & 0.058 & 72.4 & -0.581 & -0.517 & 3.076 & 57.387 & 0.193 & 0.076 & 78.0  \\
        6 & -1.392  & -1.093  & 4.356  & 48.446  & 0.145 & -0.025 & 65.3 & -0.745 & -0.549 & 2.811 & 58.260 & 0.136 & 0.004 & 74.6\\
        7 & -1.082  & -0.947  & 4.795  & 37.655  & 0.164 & -0.022 & 65.3 & -0.565 & -0.543 & 3.130 & 44.151 & 0.144 & 0.018 & 73.9 \\
        8 & -1.209  & -1.059  & 4.614  & 39.574  & 0.180 & -0.002 & 65.0 & -0.579 & -0.512 & 2.922 & 59.465 & 0.152 & 0.030 & 75.9 \\
        9 & -1.239  & -0.949  & 4.857  & 46.069  & 0.270 & 0.067 & 74.2 & -0.682 & -0.505 & 3.002 & 58.049 & 0.131 & -0.027 & 74.7 \\
        10 & -0.993  & -0.840  & 5.174  & 35.581  & 0.353 & 0.117 & 75.3 & -0.556 & -0.511 & 3.142 & 54.788 & 0.175 & 0.041 & 74.4 \\
        11 & -1.231  & -1.004  & 4.624  & 30.913  & 0.272 & 0.076 & 68.2 & -0.637 & -0.493 & 2.935 & 34.481 & 0.181 & 0.035 & 76.4 \\
        12 & -1.154  & -0.929  & 4.850  & 30.591  & 0.303 & 0.064 & 69.3 & -0.601 & -0.500 & 3.081 & 38.755 & 0.184 & 0.057 & 75.6 \\
        \hline
        13 & -1.230  & -1.054  & 5.124  & 52.974  & 0.284 & 0.076 & 70.2 & -0.584 & -0.498 & 3.200 & 60.767 & 0.199 & 0.073 & 76.6 \\
        14 & -1.226  & -0.978  & 5.186  & 53.150  & 0.301 & 0.092 & 72.2 & -0.667 & -0.530 & 3.083 & 63.175 & 0.145 & 0.005 & 74.9 \\
        15 &  &  & 5.076  & 46.615  & 0.298 & 0.110 & 74.2 & & & 3.208 & 55.076 & 0.200 & 0.081 & 75.6 \\
        16 &  &  & 4.847  & 47.360  & 0.198 & -0.075 & 58.9 & & & 3.260 & 60.138 & 0.247 & 0.141 & 69.8 \\
        17 &  &  & 4.578  & 31.131  & 0.066 & -0.319 & 40.9 & & & 3.109 & 56.697 & 0.138 & 0.012 & 65.6 \\
        18 &  &  & 4.576  & 28.835  & 0.071 & -0.309 & 38.4 & & & 3.150 & 55.033 & 0.162 & 0.043 & 64.2 \\
        19 &  &  & 5.024  & 36.493  & 0.256 & -0.680 & 65.6 & & & 3.242 & 49.445 & 0.223 & 0.108 & 76.3 \\
        20 &  &  & 4.760  & 36.451  & 0.151 & -0.093 & 58.4 & & & 3.205 & 54.922 & 0.209 & 0.102 & 71.3 \\
        21 &  &  & 4.829  & 35.495  & 0.187 & -0.691 & 64.0 & & & 3.258 & 47.359 & 0.230 & -0.435 & 75.4 \\
        22 &  &  & 4.946  & 34.103  & 0.231 & 0.034 & 62.9 & & & 3.253 & 52.114 & 0.231 & 0.119 & 77.1 \\
        \hline
        23 & -1.169  & -0.974  & 4.225  & 48.668  & 0.190 & 0.034 & 69.4  & -0.561 & -0.503 & 2.832 & 57.624 & 0.214 & 0.107 & 77.1 \\
        24 & -1.014  & -0.908  & 5.181  & 57.411  & 0.362 & 0.164 & 75.7 & -0.536 & -0.503 & 3.340 & 58.396 & 0.313 & 0.208 & 78.6 \\
        25 & -1.024  & -0.881  & 5.151  & 55.490  & 0.312 & 0.099 & 74.4 & -0.540 & -0.500 & 3.312 & 62.857 & 0.268 & 0.173 & 77.8 \\
        26 & -0.950  & -0.841  & 5.287  & 61.007  & 0.369 & 0.156 & 78.4 & -0.533 & -0.505 & 3.340 & 63.100 & 0.285 & 0.190 & 77.9 \\
        \hline
        27 & -1.034  & -0.834  & 4.609  & 63.988  & 0.302 & 0.159 & 83.4 & -0.558 & -0.484 & 2.828 & 58.549 & 0.211 & 0.105 & 77.0 \\
        28 & -0.767  & -0.630  & 5.499  & 75.592  & 0.578 & 0.400 & 91.7 & -0.534 & -0.495 & 3.363 & 61.016 & 0.341 & 0.238 & 79.1 \\
        29 & -0.784  & -0.612  & 5.493  & 78.550  & 0.531 & 0.358 & 89.0 & -0.539 & -0.493 & 3.347 & 62.604 & 0.302 & 0.203 & 78.1 \\
        30 & -0.671  & -0.518  & 5.625  & 74.917  & 0.646 & 0.447 & 91.5 & -0.536 & -0.499 & 3.371 & 66.276 & 0.312 & 0.211 & 78.7 \\
        \hline
        31 & -1.057  & -0.740  & 5.587  & 41.936  & 0.527 & 0.263 & 85.4 & -0.675 & -0.499 & 3.163 & 39.618 & 0.275 & 0.176 & 78.6 \\
        32 & -1.819  & -1.415  & 3.424  & 26.731  & -0.106 & -0.214 & 47.1 & -1.142 & -0.794 & 2.048 & 52.277 & -0.028 & -0.213 & 68.4 \\
        33 & -1.271  & -0.995  & 4.621  & 58.167  & 0.198 & -0.060 & 66.1 & -0.601 & -0.503 & 3.120 & 68.578 & 0.253 & 0.150 & 78.3 \\
        \hline
        34 &  &  & 6.188  & 47.724  & 0.075 & -0.106 & 66.0 & & & 3.198 & 64.808 & 0.275 & 0.184 & 74.9 \\
        35 &  &  & 5.546  & 84.858  & 0.869  & 0.653 & 96.0 & & & 3.143 & 67.367 & 0.377 & 0.312 & 79.5 \\
        \bottomrule
        \end{tabular}
    \end{adjustbox}
\end{table}

\begin{table}[]
    \centering
    \caption{The ranking scores and fine-tuning accuracy for Office-Home and TeraIncognita datasets. The numbering in the first column corresponds to a pre-trained model from Table~\ref{tab_supp:pre-trained_models}. The numbers in each subsequent column represent the scores assigned by a ranking metric to the PTMs. The last column displays the accuracy of each model after fine-tuning. Empty cells represent models for which ranking is not feasible.}
    \label{tab:full_result_officehome_terraincognita}
    \begin{adjustbox}{width=0.95\textwidth}
        \begin{tabular}{c|ccccccc|ccccccc}
        \toprule
        Model & \multicolumn{7}{c|}{Office-Home} & \multicolumn{7}{c}{TerraIncognita}\\
        \hline
        Number & LEEP & NCE & H-Score & kNN & LogME & ZooD & Acc. & LEEP & NCE & H-Score & kNN & LogME & ZooD & Acc. \\
        \hline
        1 & -1.540 & -1.311 & 41.908 & 50.614 & 0.985 & 0.075 & 67.7 & -1.531 & -1.286 & 5.559 & 23.477 & 0.301 & -0.722 & 31.0 \\
        2 & -1.355 & -1.198 & 43.973 & 53.499 & 1.029 & 0.120 & 70.6 & -1.501 & -1.338 & 5.592 & 29.018 & 0.305 & -0.721 & 35.2 \\
        3 & -1.465 & -1.263 & 41.439 & 51.501 & 0.979 & 0.076 & 69.1 & -1.519 & -1.290 & 5.491 & 23.227 & 0.292 & -0.735 & 25.5 \\
        4 & -1.457 & -1.280 & 35.695 & 47.413 & 0.941 & 0.025 & 68.7 & -1.473 & -1.266 & 4.850 & 22.977 & 0.244 & -0.815 & 23.9 \\
        5 & -1.460 & -1.271 & 37.727 & 48.186 & 0.952 & 0.036 & 69.1 & -1.573 & -1.321 & 5.119 & 22.116 & 0.251 & -0.831 & 23.0 \\
        6 & -2.243 & -1.701 & 30.175 & 44.089 & 0.887 & -0.015 & 59.0 & -1.636 & -1.327 & 4.432 & 24.368 & 0.238 & -0.881 & 17.7 \\
        7 & -1.396 & -1.327 & 40.696 & 53.520 & 0.977 & 0.083 & 66.2 & -1.440 & -1.286 & 5.097 & 24.285 & 0.250 & -0.819 & 23.8 \\
        8 & -1.713 & -1.439 & 32.911 & 45.934 & 0.902 & -0.005 & 62.8 & -1.614 & -1.373 & 4.782 & 22.793 & 0.264 & -0.811 & 29.7 \\
        9 & -1.628 & -1.143 & 40.378 & 51.252 & 1.022 & 0.106 & 72.2 & -1.610 & -1.388 & 5.124 & 25.737 & 0.299 & -0.740 & 32.8 \\
        10 & -1.229 & -1.082 & 45.309 & 45.939 & 1.094 & 0.176 & 73.6 & -1.523 & -1.383 & 5.517 & 25.909 & 0.319 & -0.720 & 24.8 \\
        11 & -1.528 & -1.174 & 36.781 & 47.708 & 1.018 & 0.100 & 72.5 & -1.563 & -1.393 & 4.474 & 26.624 & 0.272 & -0.746 & 30.3 \\
        12 & -1.320 & -1.099 & 42.086 & 48.265 & 1.070 & 0.139 & 75.9 & -1.545 & -1.466 & 4.984 & 25.561 & 0.289 & -0.720 & 30.9 \\
        \hline
        13 & -1.594 & -1.311 & 41.423 & 48.194 & 0.972 & 0.061 & 66.3 & -1.625 & -1.315 & 6.101 & 25.319 & 0.348 & -0.803 & 31.9 \\
        14 & -1.825 & -1.377 & 39.631 & 43.415 & 0.937 & 0.027 & 62.4 & -1.704 & -1.309 & 6.106 & 24.481 & 0.344 & -0.910 & 26.7 \\
        15 & & & 40.498 & 37.124 & 0.971 & -0.022 & 60.6 & & & 5.542 & 24.565 & 0.307 & -0.721 & 23.7 \\
        16 & & & 38.633 & 32.130 & 0.941 & -0.102 & 41.6 & & & 5.601 & 26.435 & 0.308 & -0.742 & 19.1 \\
        17 & & & 31.841 & 18.154 & 0.825 & -0.399 & 22.7 & & & 5.675 & 27.931 & 0.308 & -1.067 & 16.0 \\
        18 & & & 32.493 & 19.447 & 0.838 & -0.366 & 24.4 & & & 5.711 & 30.123 & 0.313 & -0.777 & 18.4 \\
        19 & & & 39.876 & 30.521 & 0.956 & -0.010 & 61.0 & & & 5.649 & 26.656 & 0.322 & -0.710 & 28.7 \\
        20 & & & 36.612 & 27.949 & 0.912 & -0.100 & 44.1 & & & 5.486 & 23.898 & 0.296 & -0.775 & 16.1 \\
        21 & & & 38.936 & 29.547 & 0.950 & -0.424 & 52.7 & & & 5.537 & 23.617 & 0.303 & -0.745 & 23.6 \\
        22 & & & 39.705 & 28.988 & 0.954 & -0.041 & 58.8 & & & 5.680 & 26.854 & 0.323 & -0.994 & 23.2 \\
        \hline
        23 & -1.680 & -1.400 & 26.787 & 45.371 & 0.895 & -0.028 & 62.3 & -1.560 & -1.311 & 3.817 & 23.495 & 0.228 & -0.846 & 26.5 \\
        24 & -1.339 & -1.194 & 44.073 & 49.205 & 1.049 & 0.097 & 71.2 & -1.487 & -1.322 & 5.527 & 25.801 & 0.309 & -0.698 & 32.5 \\
        25 & -1.294 & -1.156 & 44.683 & 56.220 & 1.055 & 0.151 & 72.7 & -1.505 & -1.335 & 5.439 & 24.983 & 0.291 & -0.718 & 27.7 \\
        26 & -1.168 & -1.081 & 46.671 & 60.344 & 1.106 & 0.199 & 74.8 & -1.487 & -1.360 & 5.510 & 26.461 & 0.302 & -0.685 & 28.8 \\
        \hline
        27 & -1.502 & -1.266 & 28.820 & 49.142 & 0.924 & 0.004 & 66.7 & -1.549 & -1.291 & 3.761 & 23.208 & 0.223 & -0.856 & 29.3 \\
        28 & -1.152 & -1.024 & 46.552 & 56.192 & 1.119 & 0.167 & 76.1 & -1.495 & -1.354 & 5.428 & 25.008 & 0.298 & -0.739 & 36.0 \\
        29 & -1.111 & -0.979 & 47.382 & 61.253 & 1.133 & 0.230 & 78.0 & -1.515 & -1.360 & 5.342 & 26.525 & 0.277 & -0.730 & 34.4 \\
        30 & -0.971 & -0.875 & 50.223 & 67.685 & 1.226 & 0.312 & 81.0 & -1.449 & -1.343 & 5.478 & 28.274 & 0.298 & -0.681 & 35.4 \\
        \hline
        31 & -1.252 & -0.859 & 47.500 & 60.458 & 1.240 & 0.306 & 84.6 & -1.579 & -1.392 & 4.934 & 29.336 & 0.303 & -0.669 & 37.3 \\
        32 & -3.896 & -2.913 & 15.908 & 9.459 & 0.755 & -0.178 & 31.9 & -1.828 & -1.400 & 19.076 & 24.408 & 0.230 & -0.939 & 26.2 \\
        33 & -1.675 & -1.295 & 37.045 & 58.928 & 1.027 & 0.107 & 71.8 & -1.548 & -1.268 & -0.153 & 26.017 & 0.247 & -0.827 & 21.3 \\
        \hline
        34 & & & 26.080 & 22.301 & 0.828 & -0.091 & 42.4 & & & 3.695 & 28.290 & 0.220 & -0.868 & 18.8 \\
        35 & & & 36.712 & 65.789 & 1.056 & 0.148 & 82.2 & & & 4.147 & 31.467 & 0.259 & -0.749 & 40.0 \\
        \bottomrule
        \end{tabular}
    \end{adjustbox}
\end{table}

\begin{table}[]
    \centering
    \caption{The ranking scores and fine-tuning accuracy for NICO dataset. The numbering in the first column corresponds to a pre-trained model from Table~\ref{tab_supp:pre-trained_models}. The numbers in each subsequent column represent the scores assigned by a ranking metric to the PTMs. The last column displays the accuracy of each model after fine-tuning. Empty cells represent models for which ranking is not feasible.}
    \label{tab:full_result_nico}
    \begin{adjustbox}{width=0.95\textwidth}
        \begin{tabular}{c|ccccccc|ccccccc}
        \toprule
        Model & \multicolumn{7}{c|}{NICO-Animal} & \multicolumn{7}{c}{NICO-Vehicle}\\
        \hline
        Number & LEEP & NCE & H-Score & kNN & LogME & ZooD & Acc. & LEEP & NCE & H-Score & kNN & LogME & ZooD & Acc.\\
        \hline
        1 & -0.501 & -0.397 & 7.767 & 86.348 & 0.512 & 0.510 & 91.0 & -0.699 & -0.651 & 6.758 & 81.043 & 0.398 & 0.363 & 86.1 \\
        2 & -0.419 & -0.340 & 7.975 & 88.823 & 0.599 & 0.602 & 92.8 & -0.624 & -0.598 & 6.928 & 84.266 & 0.467 & 0.433 & 88.1 \\
        3 & -0.455 & -0.379 & 7.789 & 87.400 & 0.527 & 0.525 & 92.0 & -0.670 & -0.637 & 6.773 & 82.466 & 0.405 & 0.376 & 86.7 \\
        4 & -0.450 & -0.376 & 7.358 & 86.748 & 0.466 & 0.455 & 91.8 & -0.692 & -0.661 & 6.387 & 80.092 & 0.370 & 0.329 & 86.4 \\
        5 & -0.479 & -0.375 & 7.472 & 85.773 & 0.483 & 0.471 & 92.0 & -0.720 & -0.679 & 6.469 & 79.118 & 0.381 & 0.340 & 86.6 \\
        6 & -0.983 & -0.629 & 6.721 & 78.237 & 0.343 & 0.326 & 83.7 & -1.109 & -0.834 & 5.718 & 72.119 & 0.242 & 0.206 & 79.2 \\
        7 & -0.460 & -0.450 & 7.748 & 84.286 & 0.519 & 0.502 & 88.7 & -0.647 & -0.660 & 6.659 & 77.803 & 0.371 & 0.336 & 83.6 \\
        8 & -0.616 & -0.508 & 6.810 & 81.108 & 0.326 & 0.318 & 86.6 & -0.792 & -0.743 & 5.959 & 76.653 & 0.268 & 0.233 & 82.5 \\
        9 & -0.646 & -0.345 & 7.814 & 79.292 & 0.600 & 0.583 & 92.1 & -0.823 & -0.600 & 6.739 & 80.821 & 0.474 & 0.427 & 88.0 \\
        10 & -0.393 & -0.318 & 8.089 & 82.033 & 0.693 & 0.664 & 92.4 & -0.578 & -0.560 & 7.016 & 77.957 & 0.547 & 0.493 & 88.0 \\
        11 & -0.598 & -0.309 & 7.797 & 80.542 & 0.681 & 0.656 & 93.5 & -0.742 & -0.569 & 6.655 & 80.318 & 0.526 & 0.477 & 89.1 \\
        12 & -0.460 & -0.277 & 8.201 & 80.414 & 0.811 & 0.798 & 95.1 & -0.644 & -0.545 & 6.963 & 78.615 & 0.593 & 0.545 & 90.3 \\
        \hline
        13 & -0.602 & -0.468 & 7.551 & 82.090 & 0.433 & 0.428 & 88.0 & -0.743 & -0.651 & 6.685 & 78.180 & 0.374 & 0.340 & 84.9 \\
        14 & -0.921 & -0.634 & 7.030 & 69.756 & 0.288 & 0.275 & 81.2 & -0.941 & -0.731 & 6.407 & 71.239 & 0.283 & 0.247 & 80.7 \\
        15 & & & 7.546 & 71.552 & 0.438 & 0.427 & 86.9 & & & 6.644 & 71.200 & 0.362 & 0.326 & 82.9 \\
        16 & & & 7.679 & 73.400 & 0.491 & 0.485 & 80.0 & & & 6.701 & 67.634 & 0.376 & 0.331 & 74.0 \\
        17 & & & 6.562 & 46.842 & 0.188 & 0.166 & 53.2 & & & 6.050 & 49.714 & 0.184 & 0.143 & 53.6 \\
        18 & & & 6.756 & 48.977 & 0.225 & 0.207 & 55.4 & & & 6.184 & 52.048 & 0.221 & 0.176 & 56.0 \\
        19 & & & 7.652 & 68.655 & 0.470 & 0.462 & 89.3 & & & 6.743 & 69.965 & 0.395 & 0.354 & 83.8 \\
        20 & & & 7.491 & 68.446 & 0.429 & 0.419 & 81.1 & & & 6.532 & 65.629 & 0.323 & 0.276 & 75.6 \\
        21 & & & 7.649 & 60.005 & 0.458 & -0.970 & 84.2 & & & 6.681 & 62.967 & 0.370 & -0.704 & 78.3 \\
        22 & & & 7.580 & 65.025 & 0.445 & 0.436 & 87.7 & & & 6.710 & 66.776 & 0.385 & 0.343 & 82.4 \\
        \hline
        23 & -0.482 & -0.391 & 6.713 & 84.406 & 0.404 & 0.391 & 90.6 & -0.688 & -0.633 & 5.748 & 77.967 & 0.324 & 0.284 & 85.9 \\
        24 & -0.346 & -0.278 & 8.081 & 89.122 & 0.666 & 0.656 & 94.3 & -0.593 & -0.573 & 7.001 & 83.783 & 0.524 & 0.479 & 89.9 \\
        25 & -0.333 & -0.255 & 8.266 & 88.655 & 0.754 & 0.757 & 95.1 & -0.563 & -0.538 & 7.122 & 86.015 & 0.559 & 0.519 & 90.1 \\
        26 & -0.305 & -0.245 & 8.383 & 89.750 & 0.832 & 0.831 & 95.9 & -0.524 & -0.514 & 7.250 & 87.605 & 0.627 & 0.582 & 91.1 \\
        \hline
        27 & -0.444 & -0.347 & 6.793 & 81.971 & 0.425 & 0.410 & 91.3 & -0.649 & -0.602 & 5.873 & 78.824 & 0.350 & 0.312 & 86.4 \\
        28 & -0.283 & -0.211 & 8.253 & 89.394 & 0.772 & 0.762 & 95.8 & -0.527 & -0.520 & 7.131 & 85.509 & 0.594 & 0.549 & 91.1 \\
        29 & -0.287 & -0.192 & 8.424 & 93.119 & 0.872 & 0.871 & 96.7 & -0.515 & -0.490 & 7.250 & 88.538 & 0.632 & 0.590 & 91.6 \\
        30 & -0.255 & -0.164 & 8.594 & 90.335 & 1.038 & 1.037 & 97.4 & -0.478 & -0.450 & 7.430 & 89.605 & 0.752 & 0.710 & 92.8 \\
        \hline
        31 & -0.521 & -0.167 & 8.407 & 84.414 & 1.086 & 1.063 & 97.5 & -0.641 & -0.439 & 7.254 & 90.010 & 0.824 & 0.774 & 94.5 \\
        32 & -1.864 & -1.317 & 4.772 & 35.264 & 0.057 & 0.031 & 62.2 & -1.801 & -1.282 & 4.525 & 41.243 & 0.044 & 0.007 & 64.4 \\
        33 & -0.393 & -0.224 & 8.673 & 93.392 & 0.819 & 0.798 & 94.6 & -0.616 & -0.511 & 6.808 & 89.564 & 0.589 & 0.534 & 90.4 \\
        \hline
        34 & & & 7.429 & 84.647 & 0.472 & 0.465 & 89.4 & & & 6.929 & 83.589 & 0.567 & 0.539 & 92.3 \\
        35 & & & 8.240 & 95.664 & 0.936 & 0.932 & 97.5 & & & 7.206 & 89.449 & 0.832 & 0.805 & 97.3 \\
        \bottomrule
        \end{tabular}
    \end{adjustbox}
\end{table}

\begin{table}[]
    \centering
        \caption{The ranking scores and fine-tuning accuracy for DomainNet dataset. The numbering in the first column corresponds to a pre-trained model from Table~\ref{tab_supp:pre-trained_models}. The numbers in each subsequent column represent the scores assigned by a ranking metric to the PTMs. The last column displays the accuracy of each model after fine-tuning. Empty cells represent models for which ranking is not feasible.}
    \label{tab:full_result_domainnet}
    \begin{adjustbox}{width=0.6\textwidth}
        \begin{tabular}{c|ccccccc}
        \toprule
        Model & \multicolumn{7}{c}{DomainNet} \\
        \hline
        Number & LEEP & NCE & H-Score & kNN & LogME & ZooD & Acc. \\
        \hline
        1 & -4.083 & -3.972 & 51.822 & 24.387 & 1.590 & 1.229 & 31.1  \\
        2 & -3.946 & -3.898 & 58.350 & 26.811 & 1.601 & 1.237 & 32.6  \\
        3 & -4.033 & -3.963 & 50.728 & 24.933 & 1.588 & 1.228 & 31.3  \\
        4 & -3.984 & -3.943 & 45.158 & 23.998 & 1.566 & 1.204 & 32.2  \\
        5 & -3.989 & -3.931 & 48.664 & 25.178 & 1.569 & 1.207 & 33.5  \\
        6 & -4.646 & -4.287 & 31.525 & 19.208 & 1.560 & 1.211 & 24.2  \\
        7 & -3.999 & -3.981 & 49.943 & 23.852 & 1.588 & 1.238 & 30.3  \\
        8 & -4.172 & -4.059 & 32.807 & 21.075 & 1.561 & 1.208 & 27.9  \\
        9 & -4.177 & -3.833 & 47.122 & 25.990 & 1.584 & 1.225 & 34.2  \\
        10 & -3.768 & -3.694 & 58.857 & 25.956 & 1.603 & 1.250 & 34.7  \\
        11 & -4.063 & -3.829 & 46.212 & 24.848 & 1.586 & 1.231 & 35.3  \\
        12 & -3.914 & -3.769 & 56.918 & 26.283 & 1.602 & 1.240 & 37.4  \\
        \hline
        13 & -4.127 & -3.965 & 50.865 & 24.040 & 1.588 & 1.225 & 31.8  \\
        14 & -4.252 & -4.037 & 48.624 & 21.554 & 1.584 & 1.224 & 30.8  \\
        15 & & & 52.079 & 20.940 & 1.591 & 1.211 & 27.1  \\
        16 & & & 54.303 & 17.481 & 1.597 & 1.179 & 12.7  \\
        17 & & & 30.438 & 8.729 & 1.556 & 1.113 & 4.1  \\
        18 & & & 33.129 & 9.266 & 1.560 & 1.117 & 4.5  \\
        19 & & & 47.827 & 17.507 & 1.584 & 1.200 & 25.4  \\
        20 & & & 48.762 & 16.188 & 1.587 & 1.174 & 15.1  \\
        21 & & & 51.271 & 15.744 & 1.591 & 1.191 & 18.5 \\
        22 & & & 47.734 & 16.392 & 1.583 & 1.203 & 23.1  \\
        \hline
        23 & -4.078 & -3.992 & 28.905 & 22.296 & 1.558 & 1.198 & 29.7  \\
        24 & -3.787 & -3.793 & 64.463 & 27.011 & 1.613 & 1.233 & 38.3  \\
        25 & -3.788 & -3.743 & 64.207 & 28.979 & 1.614 & 1.250 & 35.7  \\
        26 & -3.661 & -3.685 & 70.961 & 30.872 & 1.626 & 1.260 & 38.1  \\
        \hline
        27 & -3.841 & -3.748 & 35.255 & 27.955 & 1.569 & 1.215 & 35.9  \\
        28 & -3.426 & -3.430 & 82.151 & 35.589 & 1.648 & 1.282 & 46.3  \\
        29 & -3.413 & -3.380 & 83.818 & 38.643 & 1.654 & 1.300 & 44.7  \\
        30 & -3.229 & -3.224 & 98.610 & 42.285 & 1.687 & 1.328 & 48.2  \\
        \hline
        31 & -3.646 & -3.376 & 73.872 & 35.363 & 1.635 & 1.277 & 48.8  \\
        32 & -5.639 & -5.096 & 14.577 & 5.968 & 1.536 & 1.178 & 10.6  \\
        33 & -4.226 & -3.908 & 50.099 & 27.670 & 1.593 & 1.232 & 34.1  \\
        \hline
        34 & & & 43.703 & 16.713 & 1.565 & 1.201 & 15.9  \\
        35 & & & 54.259 & 49.147 & 1.601 & 1.259 & 56.2  \\
        \bottomrule
        \end{tabular}
    \end{adjustbox}
\end{table}

\newpage

\section{Model Ranking in ZooD}
In this section, we present more details about the proposed ranking metric and algorithm.

\subsection{Preliminaries: setup, problem and strategy}\label{App3:sec1}

Suppose that:
\begin{itemize}
    \item {\bf Model zoo.} We have a collection of PTMs as learned feature extractors: \benrr
    \cM = \{\phi_1(x), \phi_2(x), \ldots, \phi_k(x), \ldots \},
    \eenrr 
    where $\phi_k(x)$ is a $d$-dimensional feature extractor that maps $\cX$ to $\R^d$. 
    \item {\bf Dataset.} A multi-domain dateset is collected for solving a domain generalization problem: 
    \benrr
    \cD=\{\cD_1,\cD_2,\ldots, \cD_m \}, \text{ with } \cD_i=\big\{(x_{ij}, y_{ij}), 1\leq j \leq n_i\big\},
    \eenrr 
    where $m$ is the number of observed domains and $\cD_i$ is the set of data points under the $i$-th domain. The total sample size is $n = \sum_i n_i.$
    \item {\bf Problem.} The objective is to select a PTM $\phi$ from $\cM$ such that the optimal top classifier $f$ based on the selected feature extractor $\phi$, i.e. the whole predictor is $f \circ \phi (x)$, has good prediction performance on the domain generalization task. 
\end{itemize}

To proceed further, we need more notations as folllows:
\begin{itemize}
    \item For any domain $i$, we rewrite $\cD_i=\{ \rvy_i, \rvx_i\}$ where
    \benrr
    \rvy_i = (y_{i1},y_{i2}, \ldots, y_{i n_i})^\top \in \R^{n_i}, \quad \rvx_i = (x_{i1}, x_{i2}, \ldots, x_{i  n_i})^\top \in \R^{n_i\times p}.
    \eenrr
    \item Given a feature extractor $\phi$, the learned feature matrix is denoted by
    \benrr
    \Phi_i = \big( \phi(x_{i1}), \phi(x_{i2}), \ldots, \phi(x_{i n_i})  \big)^\top \in \R^{n_i \times d}.
    \eenrr
    \item For any $i \in [m]$, we denote $\Phi_{-i}$ and $\rvy_{-i}$ as
    \benrr
    \rvy_{-i} &=& \big(\rvy_1^\top, \cdots, \rvy_{i-1}^\top, \rvy_{i+1}^\top, \cdots, \rvy_m^\top\big)^\top \in \R^{(n-n_i)},  \\
    \Phi_{-i} &=& \big( \Phi_1^\top, \cdots, \Phi^\top_{i-1}, \Phi_{i+1}^\top, \cdots, \Phi_m^\top \big)^\top \in \R^{(n-n_i)\times d}.
    \eenrr
\end{itemize}

We can break the model selection problem down into two questions. 1). When generalizing to unknown domains, are the learned features stable enough to avoid extrapolating predictions? 2). Are the learned features informative enough to ensure that the correlation between features and labels is stable across different domains?
To answer these two questions, we compute the following two quantities:
\begin{itemize}
    \item $p(\Phi_i | \Phi_{-i})$, which measures covariate shift between $\Phi_i$ and $\Phi_{-i}$, indicating whether the validation input is a rare sample compared with the training input;
    \item $p(\rvy_i | \Phi_i, \rvy_{-i}, \Phi_{-i})$, which measures the discriminability and correlation shift between $\Phi_i$ and $\rvy_i$ given the training data $\Phi_{-i}$ and $\rvy_{-i}$. 
\end{itemize}
We thus propose a metric by assembling the above quantities for PTMs ranking:
\benr\label{app:metric}
\log \text{ } p(\rvy_i | \Phi_i, \rvy_{-i},\Phi_{-i}) + \lambda \log \text{ } p(\Phi_i | \Phi_{-i}),
\eenr
where $\lambda$ is a tuning parameter that unifies the scale of the correlation shift and the covariate shift. In our  implementation, the tuning parameter is taken to be the ratio of two standard deviations: 
\benrr
\lambda = \frac{\text{Std}(\log \text{ } p(y_{ij} | \Phi_i, \rvy_{-i},\Phi_{-i}))}{\text{Std}(\log \text{ } p(\phi(x_{ij}) | \Phi_{-i}))},
\eenrr 
which is also used in~\citet{ye2021towards}.

\subsection{Model Assumption}\label{App3:sec2}
Since the correlation between $\phi(x)$ and response variables $y$ may be non-linear, we need to make further assumptions and approximations. Let each $y$ be independently generated from a unknown distribution: $p(y| \Phi, f)$. Assume this distribution is unimodal and the mode is denoted by $\mu$, we can take Taylor expansion of log-likelihood at the mode
\[
\log \text{ } p\big(y| \phi(x), f\big) \approx \log \text{ } p\big(\mu \big| \phi(x), f\big) - \frac{1}{2} (y-\mu)^\top \Lambda (y-\mu)
\]
where $\Lambda= - \nabla_y \nabla_y \log p(y| \phi(x), f)\big|_{y=\mu}$. The above transformation is the Laplace approximation~\cite{MacKay1998Laplace} and the quadratic term implies the rationality of the Gaussian approximation. Similar to~\citet{you2021logme}, the top model over a learned feature extractor $\phi$ is approximated with a linear model:
\benrr
y = \rvw^\top \phi(x) + \epsilon, \quad y \in \R, \,\, \rvw \in \R^d, \,\,  \epsilon \in \R,
\eenrr
where $\epsilon$ is Gaussian noise with variance $\beta^{-1}$. We assume the prior distribution of the weights $\rvw$ is a zero-mean isotropic Gaussian distribution governed by a hyperparameter $\alpha$:
\benrr
\rvw \sim \cN(\mathbf{0}, \alpha^{-1} \bbI_d) \quad \text{or} \quad p(\rvw;\alpha) = \Big(\frac{\alpha}{2\pi}\Big)^{\frac{d}{2}}\exp\Big(-\frac{\alpha}{2}\rvw^\top \rvw\Big)
\eenrr
and the conditional distribution of the target variable $y$ given $\phi(x)$ is a Gaussian distribution:
\benrr
y\big|\phi(x), \rvw \sim \cN(\rvw^\top \phi(x), \beta^{-1}) \quad \text{or} \quad 
p\big(y \big|\phi(x), \rvw; \beta\big) = \Big(\frac{\beta}{2\pi}\Big)^{\frac{1}{2}} \exp\left(-\frac{\beta}{2}\big(y- \rvw^\top \phi(x)\big)^2\right).
\eenrr
Recall the notations $\rvy_i$, $\Phi_i$, $\rvy_{-i}$ and $\Phi_{-i}$ in Appendix~\ref{App3:sec1}. Then we have
\benrr
\rvy_i|\Phi_i, \rvw \sim \cN(\Phi_i \rvw, \beta^{-1}\bbI_{n_i}) \quad \text{and} \quad \rvy_{-i}|\Phi_{-i}, \rvw \sim \cN(\Phi_{-i} \rvw, \beta^{-1}\bbI_{n-n_i}).
\eenrr 
In the next section, we present the details of estimating the two hyperparameters $\alpha$ and $\beta.$
Appendix~\ref{App3:sec4} shows how to compute the conditional density $p(\rvy_i | \Phi_i, \rvy_{-i},\Phi_{-i})$ and $p(\Phi_i | \Phi_{-i})$ in the proposed metric (\ref{app:metric}).

\subsection{Parameter Estimation}\label{App3:sec3}

If we introduce a uniform prior distribution over $\alpha$ and $\beta$, the posterior distribution for $\alpha$ and $\beta$ is
\benrr
p(\alpha, \beta| \rvy_{-i}, \Phi_{-i}) = \frac{p(\alpha, \beta, \rvy_{-i}, \Phi_{-i}) }{p( \rvy_{-i}, \Phi_{-i} ) } \propto p(\alpha, \beta, \rvy_{-i}, \Phi_{-i}) = p(\rvy_{-i}, \Phi_{-i}| \alpha, \beta) p(\alpha, \beta),
\eenrr
where the prior distribution $ p(\alpha, \beta)$ is assumed to be a uniform distribution over $\alpha$ and $\beta$. 
Then the values of $\hat \alpha$ and $\hat \beta$ are obtained by maximizing the density function $p(\rvy_{-i}, \Phi_{-i} | \alpha, \beta)$, which is also the model evidence over $\{\rvy_{-i}, \Phi_{-i}\}.$
The density function $p(\rvy_{-i}, \Phi_{-i}|\alpha, \beta)$ is obtained by integrating over $\rvw$:
\benrr
p\big(\rvy_{-i}, \Phi_{-i} \big| \alpha, \beta \big) &=& \int_\rvw  p\big(\rvy_{-i}, \Phi_{-i} \big| \rvw, \beta \big) p\big( \rvw \big| \alpha \big) \mathrm{d} \rvw \\
&=&  \int_\rvw  p\big(\rvy_{-i} \big| \Phi_{-i}, \rvw, \beta \big) p\big( \Phi_{-i} \big| \rvw, \beta \big) p\big(\rvw \big| \alpha \big) \mathrm{d} \rvw \\
&=&  \int_\rvw  p\big(\rvy_{-i} \big| \Phi_{-i}, \rvw, \beta \big) p\big(\rvw \big| \alpha \big) \mathrm{d} \rvw \times p(\Phi_{-i}) \\
&\propto& \int_\rvw  p\big( \rvy_{-i} \big| \Phi_{-i}, \rvw, \beta \big) p\big(\rvw \big| \alpha \big) \mathrm{d} \rvw.
\eenrr
According to the model assumptions in Appendix~\ref{App3:sec2}:
\benrr
\rvy_{-i}|\Phi_{-i}, \rvw \sim \cN(\Phi_{-i} \rvw, \beta^{-1}\bbI_{n-n_i}) \quad \text{and} \quad \rvw \sim \cN(\mathbf{0}, \alpha^{-1} \bbI_d), 
\eenrr 
then the likelihood function of $\alpha$ and $\beta$ is
\benrr
L(\alpha, \beta) &=& \int_\rvw  p(\rvy_{-i}|\Phi_{-i}, \rvw, \beta) p(\rvw|\alpha) \mathrm{d} \rvw \\
&=& \Big(\frac{\beta}{2\pi}\Big)^{\frac{n-n_i}{2}} \Big(\frac{\alpha}{2\pi}\Big)^{\frac{d}{2}} \int_\rvw \exp\left( -\frac{\beta}{2} (\rvy_{-i} - \Phi_{-i} \rvw)^\top (\rvy_{-i} - \Phi_{-i} \rvw) -\frac{\alpha}{2}\rvw^\top \rvw \right) \mathrm{d} \rvw \\
&=& \Big(\frac{\beta}{2\pi}\Big)^{\frac{n-n_i}{2}} \Big(\frac{\alpha}{2\pi}\Big)^{\frac{d}{2}} \int_\rvw \exp\left( - E(\rvw) \right) \mathrm{d} \rvw,
\eenrr
where $E(\rvw)$ is the energy function of $\rvw$, i.e.
\benrr
E(\rvw) =  \frac{\beta}{2} (\rvy_{-i} - \Phi_{-i} \rvw)^\top (\rvy_{-i} - \Phi_{-i} \rvw) + \frac{\alpha}{2}\rvw^\top \rvw.
\eenrr
Given $\rvy_{-i}$ and $\Phi_{-i}$, then the posterior distribution of $\rvw$ is
\benrr
 p(\rvw | \rvy_{-i}, \Phi_{-i}, \alpha, \beta) \sim \cN\left(\rvw| \rvm_{-i}, \rmA_{-i}^{-1} \right),
\eenrr
where 
\benrr
\rvm_{-i} = \beta \rmA_{-i}^{-1} \Phi_{-i}^\top \rvy_{-i}, \quad \rmA_{-i} = \alpha \bbI_d + \beta \Phi_{-i}^\top \Phi_{-i}.
\eenrr
Notice that
\benrr
E(\rvw)&=& \frac{\beta}{2} \rvw^\top \Phi_{-i}^\top \Phi_{-i} \rvw + \frac{\alpha}{2}\rvw^\top \rvw  - \beta \rvy_{-i}^\top \Phi_{-i} \rvw + \frac{\beta}{2} \rvy_{-i}^\top \rvy_{-i} \\
&=& \frac{1}{2} \rvw^\top (\beta \Phi_{-i}^\top \Phi_{-i} + \alpha \bbI_d) \rvw -  \beta \rvy_{-i}^\top \Phi_{-i} \rvw + \frac{\beta}{2} \rvy_{-i}^\top \rvy_{-i} \\
&=& \frac{1}{2} \rvw^\top \rmA_{-i} \rvw -  \beta \rvy_{-i}^\top \Phi_{-i} \rmA_{-i}^{-1} \rmA_{-i} \rvw + \frac{\beta}{2} \rvy_{-i}^\top \rvy_{-i} \\
&=& \frac{1}{2} \rvw^\top \rmA_{-i} \rvw -  \rvm_{-i}^\top \rmA_{-i} \rvw + \frac{\beta}{2} \rvy_{-i}^\top \rvy_{-i}. 
\eenrr
Then we have
$
E(\rvm_{-i}) = -\frac{1}{2} \rvm_{-i}^\top \rmA_{-i} \rvm_{-i} + \frac{\beta}{2} \rvy_{-i}^\top \rvy_{-i}.
$
We rewrite $\rvw = \rvw - \rvm_{-i}+\rvm_{-i}$ and obtain that
\benrr
\frac{1}{2} \rvw^\top \rmA_{-i} \rvw &=& \frac{1}{2} (\rvw-\rvm_{-i})^\top \rmA_{-i} (\rvw-\rvm_{-i}) - \frac{1}{2} \rvm_{-i}^\top \rmA_{-i} \rvm_{-i} + \rvm_{-i}^\top \rmA_{-i} \rvw.
\eenrr
Therefore,
\benrr
E(\rvw) &=& \frac{1}{2} (\rvw-\rvm_{-i})^\top \rmA_{-i} (\rvw-\rvm_{-i})  - \frac{1}{2} \rvm_{-i}^\top \rmA_{-i} \rvm_{-i} + \frac{\beta}{2} \rvy_{-i}^\top \rvy_{-i} \\
&=& E(\rvm_{-i}) + \frac{1}{2} (\rvw-\rvm_{-i})^\top \rmA_{-i} (\rvw-\rvm_{-i}).
\eenrr
Then we have
\benr\label{app:evidence}
\log L(\alpha, \beta) &=& \frac{n-n_i}{2} \log \beta + \frac{d}{2} \log \alpha -  \frac{n-n_i}{2} \log (2\pi) - E(\rvm_{-i}) - \frac{1}{2} \log |\rmA_{-i}| \\
&=& \frac{n-n_i}{2} \log \beta + \frac{d}{2} \log \alpha -  \frac{n-n_i}{2} \log (2\pi) - \frac{\beta}{2} \big\|\rvy_{-i}-\Phi_{-i} \rvm_{-i}\big\|^2 - \frac{\alpha}{2}\|\rvm_{-i}\|^2 - \frac{1}{2}\log |\rmA_{-i}|. \nonumber 
\eenr
and obtain $\hat \alpha$ and $\hat \beta$ by maximizing $\log L(\alpha, \beta)$, i.e.,
\benrr
\hat \alpha, \hat \beta = \argmax_{\alpha, \beta} \text{ } \log L(\alpha, \beta).
\eenrr 
We can find that the objective function here is the same as Eq.(2) in~\citet{you2021logme}. 
Then we use the fix-point iteration algorithm~\cite{you2021logme,you2021ranking}.
The detailed inference procedure is presented as follows.

Let $\lambda_i$ and $\rvv_i$ be the $i$-th eigenvalue and eigenvector of the matrix $\beta \Phi_{-i}^\top \Phi_{-i}.$ That is 
$
(\beta \Phi_{-i}^\top \Phi_{-i}) \rvv_i = \lambda_i \rvv_i.
$
Then we have
\benrr
|\rmA_{-i}| = |\alpha \bbI_d + \beta \Phi_{-i}^\top \Phi_{-i}| = \prod_{i=1}^d (\alpha+\lambda_i). 
\eenrr
The stationary points of $\log L(\alpha, \beta)$ with respect to $\alpha$ satisfy
\benrr
&& \frac{d}{2\alpha} - \frac{1}{2} \|\rvw\|^2 - \frac{1}{2} \frac{\mathrm{d}}{\mathrm{d} \alpha} \log \left( \prod_{i=1}^d (\alpha+\lambda_i) \right) = 0 \\
&\Leftrightarrow& d - \sum_{i=1}^d \frac{\alpha}{\alpha + \lambda_i}  = \alpha \|\rvw\|^2 \\
&\Leftrightarrow& \alpha = \frac{\gamma}{\|\rvw\|^2} \quad \text{with} \quad \gamma = \sum_{i=1}^d \frac{\lambda_i}{\alpha + \lambda_i}.
\eenrr
Notice that the eigenvalues $\lambda_i$ are proportional to $\beta$.
Hence $\mathrm{d} \lambda_i / \mathrm{d} \beta = \lambda_i/\beta$. Then the stationary points of $\log L(\alpha, \beta)$ with respect to $\beta$ satisfy
\benrr
&&\frac{n-n_i}{2\beta} - \frac{1}{2}\big\|\rvy_{-i}-\Phi_{-i} \rvm_{-i}\big\|^2 -\frac{1}{2\beta} \sum_{i=1}^d \frac{\lambda_i}{\alpha + \lambda_i} =0 \\
&\Leftrightarrow& \frac{1}{\beta} = \frac{1}{n - n_i -\gamma} \big\|\rvy_{-i}-\Phi_{-i} \rvm_{-i}\big\|^2.
\eenrr

\subsection{Computing Metric}\label{App3:sec4}

In this section, we present the details of computing the covariate shift $p(\Phi_i|\Phi_{-i})$ and the correlation shift $p(\rvy_i | \Phi_i, \rvy_{-i}, \Phi_{-i}).$
Then we can plug these two quantities into (\ref{app:metric}) to compute the proposed metric.

{\bf Covariate shift.} Leaving the $i$-th domain out, we compute the density $p\big(\Phi_i\big|\Phi_{-i}\big)$ to check whether the learned feature $\phi(x)$ is stable such that the distribution shift between $\Phi_i$ and $\Phi_{-i}$ is not significant. We approximate the distribution of $\phi(x)$ with a Gaussian distribution $\cN(\mu_\phi, \Sigma_\phi)$ and empirically estimate the parameters $\mu_\phi$ and $\Sigma_\phi$ from the training inputs $\Phi_{-i}\in \R^{(n-n_i)\times d}.$ That is,
\benrr
\hat \mu_\phi = \frac{1}{n-n_i} \Phi_{-i}^\top \mathbbm{1}_{n-n_i} \quad \hat \Sigma_\phi = \frac{1}{n-n_i} (\Phi_{-i} - \mathbbm{1}_{n-n_i} \hat \mu_\phi^\top )^\top (\Phi_{-i} - \mathbbm{1}_{n-n_i} \hat \mu_\phi^\top ),
\eenrr 
where $\mathbbm{1}_{n-n_i}$ is a $(n-n_i)$-length one vector.
Then we compute the density of $\Phi_i$ according to $\cN(\hat \mu_\phi, \hat \Sigma_\phi)$:
\benrr
p(\Phi_i|\Phi_{-i}) &=& p(\Phi_i|\hat \mu_\phi, \hat \Sigma_\phi) = \prod_{j=1}^{n_i} \sqrt{\frac{1}{(2\pi)^d |\hat \Sigma_\phi| }} \exp\left(-\frac{1}{2}(\phi(x_{ij})-\hat \mu_\phi)^\top \hat \Sigma_\phi^{-1} (\phi(x_{ij})-\hat \mu_\phi) \right). \\
&=&(2\pi)^{-\frac{n_i d}{2}} |\hat \Sigma_\phi|^{-\frac{n_i}{2}} \exp\left(-\frac{1}{2}  \text{trace}\big\{ (\Phi_{i} - \mathbbm{1}_{n_i} \hat \mu_\phi^\top ) \hat \Sigma_\phi^{-1} (\Phi_{i} - \mathbbm{1}_{n_i} \hat \mu_\phi^\top )^\top  \big\}\right).
\eenrr 

{\bf Correlation shift.} 
Given $\hat \alpha$ and $\hat \beta$, we have
\benr\label{app:correlation shift}
p(\rvy_i | \Phi_i, \rvy_{-i}, \Phi_{-i}; \hat \alpha, \hat \beta) = \frac{p(\rvy_i, \rvy_{-i} | \Phi_i, \Phi_{-i}; \hat \alpha, \hat \beta)}{p(\rvy_{-i}| \Phi_i, \Phi_{-i}; \hat \alpha, \hat \beta)} = \frac{p(\rvy_i, \rvy_{-i} | \Phi_i, \Phi_{-i}; \hat \alpha, \hat \beta)}{p(\rvy_{-i} | \Phi_{-i}; \hat \alpha, \hat \beta)}.
\eenr
We write $\hat \rvm_{-i} = \hat \beta \hat\rmA_{-i}^{-1} \Phi_{-i}^\top \rvy_{-i}$ and $\hat \rmA_{-i} = \hat\alpha \bbI_d + \hat \beta \Phi_{-i}^\top \Phi_{-i}.$
According to (\ref{app:evidence}),
\benr\label{app:shift1}
\log \text{ } p(\rvy_{-i} | \Phi_{-i}; \hat \alpha, \hat \beta) &=& \frac{n-n_i}{2} \log \hat\beta + \frac{d}{2} \log \hat\alpha -  \frac{n-n_i}{2} \log (2\pi) \\
&& - \frac{\hat \beta}{2} \big\|\rvy_{-i}-\Phi_{-i} \hat\rvm_{-i}\big\|^2 - \frac{\hat \alpha}{2}\|\hat\rvm_{-i}\|^2 - \frac{1}{2}\log |\hat\rmA_{-i}|. \nonumber
\eenr
To proceed further, we denote
\benrr
\rvy = (\rvy_i^\top, \rvy_{-i}^\top  )^\top \in \R^n, \quad \Phi= (\Phi_i^\top, \Phi_{-i}^\top)^\top \in \R^{n \times d}, \quad \hat \rvm = \hat \beta \hat\rmA^{-1} \Phi^\top \rvy, \quad \hat \rmA = \hat\alpha \bbI_d + \hat \beta \Phi^\top \Phi. 
\eenrr 
Similar to (\ref{app:evidence}), we have
\benr\label{app:shift2}
 \log \text{ } p(\rvy | \Phi; \hat \alpha, \hat \beta)  &=&  \log p(\rvy_i, \rvy_{-i} | \Phi_i, \Phi_{-i}; \hat \alpha, \hat \beta) \nonumber \\ 
&=& \frac{n}{2} \log \hat\beta + \frac{d}{2} \log \hat\alpha -  \frac{n}{2} \log (2\pi) - \frac{\hat \beta}{2} \big\|\rvy-\Phi \hat\rvm\big\|^2 - \frac{\hat \alpha}{2}\|\hat\rvm\|^2 - \frac{1}{2}\log |\hat\rmA|.
\eenr 
Plugging (\ref{app:shift1}) and (\ref{app:shift2}) into (\ref{app:metric}), we obtain the value of the proposed metric.

{\bf Remark.} Given $\rvy_{-i}$, $\Phi_{-i}$, $\hat \alpha$ and $\hat \beta$, the posterior distribution of $\rvw$ is
\benrr
 p(\rvw | \rvy_{-i}, \Phi_{-i}, \hat \alpha, \hat \beta) \sim \cN\left(\rvw| \hat \rvm_{-i}, \hat \rmA_{-i}^{-1} \right).
\eenrr
Further, 
\benrr
p(\rvy_i | \Phi_i, \rvy_{-i}, \Phi_{-i}; \hat \alpha, \hat \beta)
=\int_\rvw  p( \rvy_i | \Phi_i, \rvw; \hat \beta) p(\rvw | \rvy_{-i}, \Phi_{-i}; \hat \alpha, \hat \beta) \mathrm{d} \rvw.
\eenrr
By calculating the integral, we can deduce
\benrr
\rvy_i \big| \Phi_i, \rvy_{-i}, \Phi_{-i} \sim \cN\big(\Phi_i \hat\rvm_{-i}, \hat\beta^{-1} \bbI_{n_i} + \Phi_i \hat\rmA_{-i}^{-1} \Phi_i^\top \big).
\eenrr
Therefore we can also use this distribution to calculate $p(\rvy_i | \Phi_i, \rvy_{-i}, \Phi_{-i})$ directly. 
Throughout this paper, we use the formula (\ref{app:correlation shift}) to calculate the correlation shift.

\subsection{Cross-Domain Validation Selects Invariant Features}\label{app:cross}
\newcommand{\rbr}[1]{\left(#1\right)}
To justify our proposed selection method, and provide more intuition, we conduct explicit analysis in a linear regression setting. Despite the over-simplification, it does reflect the essence of our approach. From this base case, adaptions to more complicated and realistic assumptions can be made. 

\paragraph{Data Assumption} 
Suppose we have data in different domains with domain invariant and domain-specific features, with respect to the response variable $y$. Denote the set of invariant features to be $iv$, which are assumed to be unit-norm and orthogonal to each other.  
Without loss of generality, let data in domain $\mathcal{D}$ be $\bm{x} = (\bm{x}_{iv}, \bm{x}_{\mathcal{D}})$ where $\bm{x}_{iv}\in\mathbb{R}^{d^*}$ denotes the domain invariant features and $\bm{x}_{\mathcal{D}}\in\mathbb{R}^{d-d^*}$ denotes domain specific ones. Let $\bm{x}_{iv}$ be fixed. 
The domain-specific features can have non-zero correlation with $\bm{x}_{iv}$ such that 
\[\bm{x}_{\mathcal{D}} = \bm{x}_{iv}\cdot \bm{A}_{\mathcal{D}} + \bm{e}_{\mathcal{D}},\]
where $\bm{A}_{\mathcal{D}}\in\mathbb{R}^{d^*\times (d-d^*)}$, and $\bm{e}_{\mathcal{D}}\sim N(0, s^2\bm{I}_{d-d^*})$. For different domains, assume the correlation to be independently random, i.e., $\bm{A}_{\mathcal{D}}$'s are i.i.d. matrices with independent entries with mean 0 and variance 1. 
Given the features $\bm{x}$, assume the response $y$ only depends on $\bm{x}_{iv}$ such that
\[
y = \bm{x}_{iv}\cdot \beta_{iv} + \epsilon = \bm{x}\cdot \beta + \epsilon,
\]
where $\beta = (\beta_{iv}, \beta^{\mathcal{D}})$ with $\beta^{\mathcal{D}}=\mathbf{0}$ and $\epsilon$ follows $N(0,\sigma^2)$.

\paragraph{Model Assumption} 
Let the model candidates be linear models fitted to different subsets of the features and there are in total $2^{d}$ different combinations. Denote the fitted parameters to be $\hat{\beta}\in\mathbb{R}^d$ with only the selected dimensions being non-zero. Let the selection be $\phi$, which is a subset of $\{1,\ldots, d\}$. 
We want to show that our proposed statistics, in the cross-validated fashion, will prefer the optimal one with $\phi = iv$. 
The optimality is in the sense that it achieves the best goodness-of-fit, measured by the square loss. 

\paragraph{More Notations}
Let $(\bm{X}, \bm{y}), (\tilde{\bm{X}}, \tilde{\bm{y}})$ be independent datasets in two domains to be cross validated. 
For any vector (matrix), we use subscript to denote part of it with selected rows (columns). For instance, a model candidates with feature dimensions $\phi$ will only fit $\bm{y}\sim\bm{X}_\phi$ and the resulting $\hat{\beta}$ will only be nonzero on $\hat{\beta}_\phi$. For a set $\phi$, denote $|\phi|$ be to its cardinality and $\bar{\phi}$ to be its complement.

In our proposed test statistics, there are two terms to be assessed. 
The first term is essentially the goodness-of-fit of $\tilde{\bm{y}}$ and $\tilde{\bm{X}}_\phi\cdot \hat{\beta}_\phi$, which is of critical importance for selecting the invariance and consistent features across different domains. The second term can be seen as some regularization. 
In this section, we will focus on the first term, and to make things really simple, we consider expected $l_2$ loss as the measure for goodness-of-fit.

The estimated $\hat{\beta}$ can be explicitly written as
\begin{align*}
    \hat{\beta}_\phi = (\bm{X}_\phi^\top \bm{X}_\phi)^{-1} \bm{X}_\phi^\top \bm{y}\in\mathbb{R}^{|\phi|}.
\end{align*}
Given $\bm{y} = \bm{X}\beta + \bm{\epsilon}$, we can write
\[
    \bm{X}\beta = \bm{X}_\phi\beta_\phi +  \bm{X}_{\bar{\phi}}\beta_{\bar{\phi}}.
\]
Thus, 
\begin{align}
\label{beta}
    \hat{\beta}_\phi &= \beta_\phi + (\bm{X}_\phi^\top \bm{X}_\phi)^{-1} \bm{X}_\phi^\top \bm{X}_{\bar{\phi}}\beta_{\bar{\phi}} + (\bm{X}_\phi^\top \bm{X}_\phi)^{-1} \bm{X}_\phi^\top\bm{\epsilon} \nonumber \\ 
    &= \beta_\phi + (\bm{X}_\phi^\top \bm{X}_\phi)^{-1} \bm{X}_\phi^\top\bm{\epsilon}.
\end{align}
The expected $l_2$ loss can be expressed as 
\begin{align*}
   & \mathbb{E}_{\epsilon, \tilde{\epsilon}, e, A, \tilde{e},  \tilde{A}} \rbr{\|\tilde{\bm{y}}-\tilde{\bm{X}}_\phi\cdot \hat{\beta}_\phi\|^2}\\
   =& \mathbb{E}_{\epsilon, e, A, \tilde{e},  \tilde{A}} \rbr{\|\tilde{\bm{X}} \cdot\beta -\tilde{\bm{X}}_\phi\cdot \hat{\beta}_\phi\|^2} + n\sigma^2\\
   =& \mathbb{E}_{\epsilon, e, A, \tilde{e},  \tilde{A}} \rbr{\|\tilde{\bm{X}}_{iv\cap\phi}\beta_{iv\cap\phi} + \tilde{\bm{X}}_{iv\backslash\phi}\beta_{iv\backslash\phi} -\tilde{\bm{X}}_{\phi\cap iv}\cdot \hat{\beta}_{\phi\cap iv} -\tilde{\bm{X}}_{\phi\backslash iv}\cdot \hat{\beta}_{\phi\backslash iv}\|^2}+ n\sigma^2\\
   =& \mathbb{E}_{\epsilon, e, A, \tilde{e},  \tilde{A}} \rbr{\|\tilde{\bm{X}}_{iv\cap\phi}(\beta_{iv\cap\phi} - \hat{\beta}_{iv\cap\phi}) + \tilde{\bm{X}}_{iv\backslash\phi}\beta_{iv\backslash\phi}  -\tilde{\bm{X}}_{\phi\backslash iv}\cdot \hat{\beta}_{\phi\backslash iv}\|^2}+ n\sigma^2\\
   =& \mathbb{E}_{\epsilon, e, A, \tilde{e}, \tilde{A}} \rbr{\|\tilde{\bm{X}}_{iv\cap\phi} \rbr{(\bm{X}_\phi^\top \bm{X}_\phi)^{-1} \bm{X}_\phi^\top\bm{\epsilon}}_{iv\cap \phi} + \tilde{\bm{X}}_{iv\backslash\phi}\beta_{iv\backslash\phi}  -\tilde{\bm{X}}_{\phi\backslash iv}\cdot \hat{\beta}_{\phi\backslash iv}\|^2}+ n\sigma^2\\
   :=& \mathbb{E}_{\epsilon, e, A, \tilde{e},  \tilde{A}}\rbr{\|I_1 + I_2 + I_{3}\|^2} + n\sigma^2.
\end{align*}
$I_1$ accounts for the variance in estimating the selected invariance features. $I_2$ is non-random and accounts for the error from unselected invariance features. $I_3$ accounts the error from wrongly selected features. 
Easy to verify that $\mathbb{E}(I_1) = \mathbb{E}(I_3) = 0$ and $\mathbb{E}(I_1I_3)=0$, since $\hat{\beta}$ is independent with $\tilde{A},\tilde{e}$, which are both mean zero.
\begin{align*}
    \mathbb{E}_{\epsilon, e, A, \tilde{e},  \tilde{A}}(\|I_1\|^2)
    = \sigma^2 \mathbb{E}_{e, A}\mathrm{tr}\rbr{(\bm{X}_\phi^\top \bm{X}_\phi)_{iv\cap \phi}^{-1}}
\end{align*}
For $I_3$, we can further write
\begin{align*}
    \mathbb{E}_{\epsilon, e, A, \tilde{e}, \tilde{A}}(\|I_3\|^2)
     &= \mathbb{E}_{\epsilon, e, A, \tilde{e}, \tilde{A}}\rbr{\hat{\beta}_{\phi\backslash iv}^\top \tilde{\bm{X}}_{\phi\backslash iv}^\top \tilde{\bm{X}}_{\phi\backslash iv}\cdot \hat{\beta}_{\phi\backslash iv}}\\
     &=\mathbb{E}_{\epsilon, e, A}\rbr{\|\hat{\beta}_{\phi\backslash iv} \|^2} 
     \mathbb{E}_{\tilde{e}, \tilde{A}}\mathrm{tr}\rbr{\tilde{\bm{X}}_{\phi\backslash iv}^\top \tilde{\bm{X}}_{\phi\backslash iv}}\\
     &= \mathbb{E}_{\epsilon, e, A}\rbr{\|\hat{\beta}_{\phi\backslash iv} \|^2} 
     \rbr{\mathbb{E}_{\tilde{A}}\mathrm{tr}\rbr{\tilde{\bm{A}}_{\phi\backslash iv}^\top \tilde{\bm{A}}_{\phi\backslash iv}} + n|\phi\backslash iv|s^2}\\
     &= {n(1+s^2)|\phi\backslash iv|}\cdot \mathbb{E}_{\epsilon, e, A}\rbr{\|\hat{\beta}_{\phi\backslash iv} \|^2} 
\end{align*}
Therefore,
\begin{align*}
   & \mathbb{E}_{\epsilon, \tilde{\epsilon}, e, A, \tilde{e},  \tilde{A}} \rbr{\|\tilde{\bm{y}}-\tilde{\bm{X}}_\phi\cdot \hat{\beta}\|^2}\\
   =& \sigma^2 \mathbb{E}_{e, A}\mathrm{tr}\rbr{(\bm{X}_\phi^\top \bm{X}_\phi)_{iv\cap \phi}^{-1}} +\|\beta_{iv\backslash \phi}\|^2 + {n(1+s^2)|\phi\backslash iv|}\cdot\mathbb{E}_{\epsilon, e, A}\rbr{\|\hat{\beta}_{\phi\backslash iv} \|^2} + n\sigma^2.
\end{align*}
If $\phi=iv$, the above quantity is minimized with $ \mathbb{E}_{\epsilon, \tilde{\epsilon}, e, A, \tilde{e},  \tilde{A}} \rbr{\|\tilde{\bm{y}}-\tilde{\bm{X}}_\phi\cdot \hat{\beta}\|^2}=(n+d^*)\sigma^2$.

\newpage
\section{Feature Selection in ZooD}
In this section, we present more details about the PTMs ensemble and feature selection in Section~\ref{sec32}. The top-ranked PTMs in Section~\ref{sec31} are preferred for solving the OoD generalization task.  
To further aggregate different PTMs, we consider assembling the features by using PTMs as feature extractors
\[
\Phi = \big[ \Phi^{(1)}, \ldots, \Phi^{(k)}\big],
\]
where $\Phi^{(i)}$ is the $i$-th ranked feature extractor and $\left[\cdot\right]$ denotes the row concatenation operation.
As we show in experiments, in most cases, using aggregated models can significantly outperform any single model. However, the rough ensemble will inevitably introduce more noise. According to the definition of OoD learnability proposed by~\citet{ye2021towards},
non-informative but invariant features from training domains may only bring some noise, and the accumulation of noise hurts learnability of the OoD generalization task.
Therefore, we propose a Bayesian feature selection method based on the Gaussian linear framework in Section~\ref{sec31}.

\subsection{Bayesian Variable Selection}\label{FS:sec1}
In the Bayesian literature, the variable selection problem can be efficiently solved by introducing, for each variable $\rw_i$, a binary mask $\rz_i \in\{0,1\}$ \cite{liang2008mixtures,casella2009consistency,xu2015bayesian,yang2016computational}, which are given by Bernoulli distributions governed by probability coefficient $\boldsymbol{\pi}$. Let $\rz=\left\{\rz_{i}\right\}^{d}_{i=1}$ and 
\begin{equation*}
    p(\rz;\boldsymbol{\pi})=\prod_{i=1}^{d} p(\rz_{i})=\prod_{i=1}^{d} \pi_{i}^{\rz_{i}}(1-\pi_{i})^{1-\rz_{i}}.
\end{equation*}
From a generative perspective, these masks determine whether the weight $\rw_{i}$ is generated from a slab or a spike prior~\cite{ishwaran2005spike}. If $\rz_{i}=1$, then $\rw_{i}$ will follow a slab prior with diffusing probability density; if $\rz_{i}=0$, $\rw_{i}$ will have a spike prior with probability mass concentrated around 0, and thus should be discarded. Specifically, we assume
\benrr
p(\rw_i|\rz_i,\alpha_{i,1},\alpha_{i,2})=\left\{\begin{array}{ll}
\cN(0, \alpha_{i,1}^{-1}) & \text { if } \rz_i=1; \\
\cN(0, \alpha_{i,2}^{-1}) & \text { if } \rz_i=0.
\end{array}\right.
\eenrr
Denote $\rvw = (\rw_1, \ldots, \rw_d)^\top$ and $\alpha_{i,1}$ and $\alpha_{i,2}$ control the shape of the $\rw_i$ distribution and should be reasonably large for $\alpha_{i,2}$. Conditioned on $\rw_{i}$, each data point $y_{n}$ is assumed to be independently
drawn from a linear model with mean $\rvw^\top \phi(x)$ and additional Gaussian noise with inverse variance $\beta$:
\begin{equation*}
  p(y_{n} \big|\phi(x_n), \rvw; \beta) = \Big(\frac{\beta}{2\pi}\Big)^{\frac{1}{2}} \exp\left(-\frac{\beta}{2}\big(y_{n} - \rvw^\top \phi(x_{n})\big)^2\right). 
\end{equation*}
The model specification is completed by introducing conjugate Gamma priors over the inverse variance  $\beta$ and $\{\alpha_{i, 1},\alpha_{i, 2}\}^{d}_{i=1}$:
\benrr
\alpha_{i, 1} \sim \text{Gamma}(\nu_{i,1}, \nu_{i,2}), \quad \alpha_{i, 2} \sim \text{Gamma}(\nu_{i,3}, \nu_{i,4}), \quad \beta \sim \text{Gamma}(\nu_{0,1}, \nu_{0,2}).
\eenrr
Denote the set of Gamma prior parameters as $\vnu=\{\nu_{i,j}\}$ and all latent variables as
\[
\boldsymbol{\xi}=\left\{\beta, \{\rw_i,\rz_i, \alpha_{i,1}, \alpha_{i,2}\}_{i=1}^d \right\}.
\]
Then the variable selection problem can be solved by estimating $\vpi=\left\{\pi_1,\pi_2,\ldots,\pi_d \right\}$ with $\pi_i= p(\rz_i = 1).$ We can find the maximum likelihood estimator of the probability coefficient $\boldsymbol{\pi}$ of Bernoulli masks and then screen the variables if $\pi_{i}$ is smaller than the pre-defined threshold $\tau$.

\subsection{Variational EM Algorithm}\label{FS:sec2}
Given the dataset $\{\rvy,\Phi\}$, the maximum marginal likelihood
estimator of $(\vpi, \vnu)$ is given by
\benr\label{FS:eq11}
\hat \vpi, \hat \vnu &=& \argmax_{\vpi,\vnu} \text{ } \log \text{ } p(\rvy|\Phi; \vpi,\vnu) \nonumber \\
&=& \argmax_{\vpi, \vnu} \text{ } \log \int_{\vxi} p(\rvy, \vxi|\Phi; \vpi,\vnu) \mathrm{d} \vxi.
\eenr 
However, direct maximization of (\ref{FS:eq11}) is intractable due to the integration over $\vxi$. EM algorithm~\cite{rovckova2014emvs} might be a solution here. In the E-step, we compute the conditional expectation
\benrr
\cL(\vpi,\vnu; \vpi^{old},\vnu^{old}) 
&=& \E_{\vxi} \big[ \log p(\rvy, \vxi|\Phi; \vpi,\vnu) \big| \rvy,\Phi; \vpi^{old},\vnu^{old} \big] \\
&=& \int  \log p(\rvy, \vxi|\Phi; \vpi,\vnu)  p( \vxi|\rvy,\Phi; \vpi^{old},\vnu^{old}) \mathrm{d} \vxi,
\eenrr
which involves inferring posterior $p( \vxi|\rvy,\Phi; \vpi,\vnu)$. However, this is not straightforward to obtain due to the complexity of our model setup. MCMC~\cite{neal2011mcmc} is a common tool for this problem, but suffers from intensive computation, thus hard to extend to large-scale data. We instead use approximate Bayesian inference in Section~\ref{FS:sec3}.

In the M-step, we update $\vpi$ and $\vnu$ by maximizing the expectation
\[
\vpi^{new}, \vnu^{new} = \argmax_{\vpi,\vnu} \text{ } \cL(\vpi,\vnu; \vpi^{old},\vnu^{old}).  
\]
By repeating the E and M steps, the estimator $(\vpi^{new}, \vnu^{new})$ converges to an optimal solution. We show this method has satisfying performance for the underlying variable selection problems in synthetic data and the prevailing OoD dataset.

\subsection{Variational Inference}\label{FS:sec3}
In the E-Step, computation of $\E_{\vxi} \big[ \log p(\rvy, \vxi|\Phi; \vpi,\vnu) \big|\rvy,\Phi; \vpi^{old},\vnu^{old} \big]$ involves inferring posterior $p( \vxi|\rvy,\Phi; \vpi,\vnu)$. However, due to the complexity of our model setup, no analytical form of the posterior distribution can be found. We instead approximate true posterior distribution by variational inference~\cite{blei2017variational}. The main idea involves the introduction of a set of distributions $Q$, which should ideally be easy to compute and provide a good approximation to the true posterior distribution. We consider the following transformation of the marginal likelihood
\begin{equation*}
    \begin{aligned}
\ln p(\rvy|\Phi; \vpi,\vnu) &=\ln \int p(\rvy, \vxi|\Phi; \vpi,\vnu) d \vxi\\
&=\ln \int Q(\vxi) \frac{p(\rvy, \vxi|\Phi; \vpi,\vnu)}{ Q(\vxi)} d \vxi\\
&\geq \int Q(\vxi) \ln \frac{p(\rvy, \vxi|\Phi; \vpi,\vnu)}{ Q(\vxi) } d \boldsymbol{\theta} \\
&=\mathcal{L}(Q),
\end{aligned}
\end{equation*}
where $\mathcal{L}(Q)$ denotes the variational lower bound. The key point is that, through proper choice of $Q$ distribution, $\mathcal{L}(Q)$ can be readily evaluated, and thus by maximizing the lower bound, we generally find the $Q$ distribution, which is the best approximation within the considered family. Here we factorize $Q$ over each latent variable, such that
\begin{equation*}
 Q(\vxi;\vpi,\vnu)=Q(\beta;\tilde{\nu}_{0,1},\tilde{\nu}_{0,2})\prod^{d}_{i=1}\Big[Q(\rz_{i};\tilde{\pi}_{i})Q(\rw_i ;m_i,\lambda^{-1}_i)Q(\alpha_{i, 1};\tilde{\nu}_{i,1},\tilde{\nu}_{i,2})Q(\alpha_{i, 2};\tilde{\nu}_{i,3},\tilde{\nu}_{i,4})\Big],
\end{equation*}
which holds for classic mean-field family~\cite{bishop2006pattern}. By denoting $\{\boldsymbol{m},\boldsymbol{\lambda},\boldsymbol{\tilde{\pi}}\}=\{m_{i},\lambda_{i},\pi_{i}\}^{d}_{i=1}$ and $\tilde{\vnu}=\{\tilde{\nu}_{i,j}\}$, an optimization-free form over all possible $Q$ has been established, which can lead to minimization of KL divergence between variational distribution $ Q(\vxi)$ and true posterior $p(\vxi|\rvy,\Phi;\vpi,\vnu)$
\begin{equation*}
    Q^{*}\left(\vxi_{k}\right)=\frac{\exp \E_{\vxi_{-k}\sim Q^*(\vxi_{-k})}\ln p(\rvy,\vxi|\Phi;\vpi,\vnu)}{\int \exp \mathbb{E}_{\vxi_{-k}\sim Q^*(\vxi_{-k})} \ln p(\rvy,\vxi|\Phi;\vpi,\vnu)d \vxi_{k}},
\end{equation*}
where denote $\vxi_{k}$ as the $k$-th variable in the set $\boldsymbol{\xi}$ and $\vxi_{-k}$ is the subset of all other variables except $\vxi_{k}$. For models in conjugate families, the optimal $Q^{*}(\vxi_k)$ has the same form as its prior distribution. We then establish the optimization step for arbitrary variational parameters set $\{\boldsymbol{m},\boldsymbol{\lambda},\tilde{\vnu},\boldsymbol{\tilde{\pi}}\}$ to approach the true posterior:
\begin{equation*}
    m_{i}=f_{m}(\tilde{\pi}_{i},\boldsymbol{m},\tilde{\vnu})=\left(\sum_{n=1}^{N} x_{n, i}^{2} \mathbb{E}[\beta]+\tilde{\pi}_{i} \mathbb{E}[\alpha_{i, 1}]+\left(1-\tilde{\pi}_{i}\right)\mathbb{E}[\alpha_{i, 2}]\right)^{-1} \cdot \left[\mathbb{E}[\beta] \cdot \sum_{n=1}^{N} x_{n, i}\left(\sum_{j \neq i}^{d-1} m_{j} \cdot x_{n, j}-y_{n}\right)\right],
\end{equation*}
\begin{equation*}
    \tilde{\pi}_{i}=f_{\pi_{i}}(\boldsymbol{m},\boldsymbol{\lambda},\tilde{\vnu})= \frac{\exp \left\{\mathbb{E}\ln |\alpha_{i,1}| -\frac{1}{2} \mathbf{Tr}\left(\mathbb{E}[\alpha_{i, 1}] \cdot [\mathbb{E}[\rw_{i}^{2}]]\right) +\ln \pi_{i} \right\}}{\exp \left\{ \mathbb{E}\ln |\alpha_{i,1}|+ \mathbb{E}\ln |\alpha_{i,2}| -\frac{1}{2} \mathbf{Tr}\left[\left( \mathbb{E}[\alpha_{i, 1}]+ \mathbb{E}[\alpha_{i, 2}]\right) \cdot [\mathbb{E}[\rw_{i}^{2}]]\right] + \ln \pi_{i}+ \ln (1-\pi_{i}) \right\}},
\end{equation*}
\begin{equation*}
    (\tilde{\nu}_{0,2})^{-1}=f_{\nu_{0,2}}(\boldsymbol{m},\boldsymbol{\lambda})=\sum_{n=1}^{N}y_{n}^{2}-2 \sum_{n=1}^{N}\left(\sum_{i=1}^{d} m_{i} \cdot x_{n, i}\right) \cdot y_{n} +\sum_{n=1}^{N} \sum_{i, j}^{d^2} x_{n,i}\cdot x_{n j} \left([\mathbb{E}[\rw_{i}^{2}]\right) +{\nu}_{0,2}^{-1},
\end{equation*}  
\begin{equation*}
    (\tilde{\nu}_{i,2})^{-1}=f_{{\nu}_{i,2}}(\boldsymbol{m},\boldsymbol{\lambda},\tilde{\vpi})=\left(\mathbb{E}[\rw_{i}^{2}]^{-1}\right)\cdot \tilde{\pi_{i}}+{\nu}_{i,2}^{-1}, \quad
    (\tilde{\nu}_{i,4})^{-1}=f_{\nu_{i,4}}(\boldsymbol{m},\boldsymbol{\lambda},\tilde{\vpi})=\left(\mathbb{E}[\rw_{i}^{2}]^{-1}\right)\cdot(1- \tilde{\pi_{i}})+{\nu}_{i,4}^{-1},
\end{equation*}  
\begin{equation*}
    \lambda_{i}=f_{\lambda}(\tilde{\vnu})=\sum_{n=1}^{N} x_{n, i}^{2} \mathbb{E}[\beta]+\tilde{\pi}_{i} \mathbb{E}[\alpha_{i, 1}]+\left(1-\tilde{\pi}_{i}\right) \mathbb{E}[\alpha_{i, 2}],
\end{equation*}
\begin{equation*}
    \tilde{\nu}_{0,1}=f_{{\nu}_{0,1}}(n)={\nu}_{0,1}+n,
    \quad
    \tilde{\nu}_{i,1}=f_{{\nu}_{i,1}}(\tilde{\vpi})={\nu}_{i,1}+\tilde{\pi}_{i}, \quad
    \tilde{\nu}_{i,3}=f_{{\nu}_{i,3}}(\tilde{\vpi})={\nu}_{i,3}+1-\tilde{\pi}_{i},
\end{equation*}    
where the variational expectations are given by
\begin{equation}\label{Eq14}
     \mathbb{E}[\rw_{i}^{2}]= m_{i}^{2}+\lambda_{i}^{-1}, \quad 
    \mathbb{E}[\beta]= \tilde{\nu}_{0,1} \cdot  \tilde{\nu}_{0,2}, \quad 
    \mathbb{E}\left[\alpha_{i,1}\right]= \tilde{\nu}_{i,1} \cdot  \tilde{\nu}_{i,2},
    \quad
     \mathbb{E}\left[\alpha_{i,2}\right]= \tilde{\nu}_{i,3} \cdot  \tilde{\nu}_{i,4},
\end{equation}
\begin{equation}\label{Eq15}
    \mathbb{E}\ln |\alpha_{i,1}| = \psi\left(\frac{\nu^{i,1}}{2}\right)+\ln 2+\ln \left|\nu^{i,2}\right|,
    \quad
    \mathbb{E}\ln |\alpha_{i,2}| = \psi\left(\frac{\nu^{i,3}}{2}\right)+\ln 2+\ln \left|\nu^{i,4}\right|.
\end{equation}
Since the optimization steps for each variational parameter are mutually dependent, we can use coordinate gradient descent~\cite{blei2017variational} starting by current $Q(\boldsymbol{\xi})^{t-1}$ from the last iteration. After one-step optimization, variational parameters of $Q(\boldsymbol{\xi})^{t}$ are used in computation of $\E_{\boldsymbol{\xi} \sim Q(\boldsymbol{\xi};\vpi^{old},\vnu^{old})^{t}} \big[ \log p(\rvy|\Phi,\boldsymbol{\xi}; \vpi,\vnu)\big]$, thus finishing E-step. During this procedure, the lower bound $\mathcal{L}(Q)$ will continuously increase until reaching its maximum value. Therefore, the value of $\mathcal{L}(Q)$ can be used as a useful indicator for convergence of algorithm~\cite{corduneanu2001variational}.

\subsection{Algorithm Details}\label{FS:sec4}
The proposed model contains a set of prior hyper-parameters $\vpi, \vnu$, which is exactly what we want to estimate for feature screening. In Bayesian literature, hyper-parameter selection can be automated from data through a procedure named ``ARD''~\cite{mackay1995probable}. The original ``ARD'' procedure proposes a selection based on the value of model evidence. However, in many cases including ours, this evidence is intractable. Fortunately, it's also feasible to use variational lower bound $\mathcal{L}(Q)$ as a substitute. Learning prior hyper-parameters $\vpi,\vnu$ leads to the minimization of KL divergence. This can be rationalized by the decomposition of $\mathcal{L}(Q)$:
\benrr
    \mathcal{L}(Q) &=& \E_{\vxi \sim Q(\vxi)} \big[ \log \text{ }p(\rvy|\Phi, \vxi; \vpi,\vnu)\big] \\
    &=& \E_{\vxi \sim Q(\vxi)} \big[ \log \text{ } p(\rvy| \vxi,\Phi)\big] - \mathrm{KL}(Q(\vxi)||p(\vxi;\vpi,\vnu)).
\eenrr
Thus by setting derivatives of each hyper-parameters with respect to $\mathcal{L}(Q)$ to 0, it's easy to see $\mathcal{L}(Q)$ is maximized when all hyper-parameters are set to posterior parameters:
\begin{equation*}
    \vpi^{new}=\tilde{\vpi},
    \quad
    \vnu^{new}=\tilde{\vnu}.
\end{equation*}
However, the proposed algorithm still suffers from heavy computational cost: Each iteration costs $\mathcal{O}(nd^{2})$. Thus to relieve computation burden and memory usage,  we leverage our method with stochastic approximation leading to the EM algorithm with stochastic variational inference~\cite{hoffman2013stochastic}. In each iteration, we sample a random subset of entire data with size $n^{s}$. Fitting our algorithm over this subset for the current iteration, we obtain a local optimal estimator denoted by $Q^{s}(\vxi)$. In M-step these intermediate variational distributions by factorizing $Q^{s}(\vxi)$ will be used to learn hyper-parameters $\vpi$ and $\vnu$ and simultaneously as the starting point for subsequent estimator in the next iteration. In the end, we successfully reduce the computation cost to $\mathcal{O}(n^{s}d^{2})$ with $n^{s} \ll n$, while maintaining the guarantee of convergence to the global optimum~\cite{robbins1951stochastic}. In our experiments, we collect variational probabilities of $\{\tilde{\pi}_{i}\}_{i=1}^{d}$ from the last three runs and early-stop the algorithm if its difference with the current probability is smaller than the pre-defined threshold $\epsilon$ or reaches the maximum iteration times.
Variational EM algorithm for Bayesian feature selection is summarized in Algorithm~\ref{algo:feature_selection}. Note that we initialize $\boldsymbol{m}$ by linear regression and the initialization of $\tilde{\vnu}$ is set to $\vnu$.

In our experiments, we often deal with the multivariate case. If the underlying task involves multivariate regression or classification, i.e., $\boldsymbol{Y} \in \mathbb{R}^{n \times K}$, we can run the proposed EM algorithm on each dimension and take the union of all selected features. Therefore, our feature selection procedure can be used in almost all prevailing models and tasks.

\begin{algorithm}[t]
    \caption{Variational EM Algorithm for Bayesian Feature Selection}
    \label{algo:feature_selection}
    \begin{algorithmic}[1]
        \REQUIRE 
            The observed data $\boldsymbol{Y}\in \mathbb{R}^{n},\boldsymbol{X} \in \mathbb{R}^{n \times d}$; 
            Prior parameters $\vpi^0=\{\pi^0_{i}\}^d_{i=1}$ and $\vnu^{0}=\{\nu_{i,j}^{0}\}$; 
            Maximum iteration step ${T}$; 
            Batch size ${n^{s}}$; 
            Stopping threshold ${\epsilon}$.
        \ENSURE Converged $\vpi^{t}$ and $\vnu^{t}$.
        \STATE Initialization of variational moment: $\{\boldsymbol{m},\boldsymbol{\lambda},\mathbb{E}[\alpha_{i,1}],\mathbb{E}[\alpha_{i,2}],\mathbb{E}\ln |\alpha_{i,1}|,\mathbb{E}\ln |\alpha_{i,2}|\}^{d}_{i=1}$:
        \begin{itemize}
            \item Initialize $\boldsymbol{m}^{0}$ by linear regression between $\boldsymbol{Y}$ and $\boldsymbol{X}$, and let $\boldsymbol{\lambda}^{0}=(\boldsymbol{m}^{0}\odot \boldsymbol{m}^{0})^{-1}$;
           \item Set  $\tilde{\vnu}^{0}=\vnu^{0}$ and compute $E[\alpha_{i,1}],E[\alpha_{i,2}],\mathbb{E}\ln |\alpha_{i,1}|,\mathbb{E}\ln |\alpha_{i,2}|$ by Equation~\eqref{Eq14} and~\eqref{Eq15};
        \end{itemize}
        \FOR {$1 \leq t \leq T$}
        \STATE Random Sampling a data subset with size $n^{s}$;
        \STATE Update $\tilde{\nu}_{0,1}^{t}$  and $\tilde{\nu}_{0,2}^{t}$ by $f_{\nu_{0,1}^{t-1}}(n^{s})$ and $f_{\nu_{0,2}^{t-1}}(\boldsymbol{m}^{t-1},\boldsymbol{\lambda}^{t-1})$;
        \FOR {$1 \leq i \leq d$}
        \STATE Update each ${\tilde{\pi}_{i}^{t}}$ by $f_{\pi^{t-1}_{i}}(\boldsymbol{m}^{t-1},\boldsymbol{\lambda}^{t-1},\tilde{\vnu}^{t-1})$;
        \STATE Update each $\tilde{\nu}_{i,1}^{t}$, $\tilde{\nu}_{i,2}^{t}$, $\tilde{\nu}_{i,3}^{t}$, $\tilde{\nu}_{i,4}^{t}$ by $f_{\nu_{i,1}^{t-1}}(\tilde{\vpi}^{t})$, $f_{\nu_{i,2}^{t-1}}(\boldsymbol{m}^{t-1},\boldsymbol{\lambda}^{t-1},\tilde{\vpi}^{t-1})$, $f_{\nu_{i,3}^{t-1}}(\tilde{\vpi}^{t})$, $f_{\nu_{i,4}^{t-1}}(\boldsymbol{m}^{t-1},\boldsymbol{\lambda}^{t-1},\tilde{\vpi}^{t})$;
        
        \STATE Update $m^{t}_{i}$ and $\lambda^{t}_{i}$ by $f_{m}(\tilde{\pi}^{t}_{i},\boldsymbol{m}^{t-1},\tilde{\vnu}^{t})$ and $f_{\lambda}(\tilde{\vnu}^{t})$;
        \ENDFOR
        
        \STATE Update $ \vpi^{t}=\tilde{\vpi}^{t}, \vnu^{t}=\tilde{\vnu}^{t}$;
        \IF {$t \geq 3$}
            \STATE ${\vpi}^{mean}=(\vpi^{t-2}+\vpi^{t-1}+\vpi^{t})/3$;
            \STATE \textbf{Early Stop} if $|\vpi^{t}-{\vpi}^{mean}| <\epsilon$;
        \ENDIF
        \ENDFOR
    \end{algorithmic}
\end{algorithm}

\subsection{Theoretical Result}\label{FS:sec5}
It has been shown that our method, as well as others in Bayesian variable selection, has potentially strong selection consistency~\cite{liang2008mixtures,casella2009consistency,xu2015bayesian,yang2016computational}. Consider the following model with inverse Gamma prior:
\begin{equation}
    \label{model1}
    \begin{gathered}
y_{n} \mid\left(\phi(x_n), \rvw, \sigma^{2}\right) \sim \cN\left(\rvw\phi(x_n), \sigma^{2} I\right), \\
\rw_{i} \mid\left(\sigma^{2}, \rz_{i}=0\right) \sim \cN\left(0, \sigma^{2} \tau_{0, N}^{2}\right), \\
\rw_{i} \mid\left(\sigma^{2}, \rz_{i}=1\right) \sim \cN\left(0, \sigma^{2} \tau_{1, N}^{2}\right), \\
p\left(\rz_{i}=1\right)=1-p\left(\rz_{i}=0\right)=q_{N}, \\
\sigma^{2} \sim \mathrm{IG}\left(\alpha_{1}, \alpha_{2}\right),
\end{gathered}
\end{equation}
where $i$ runs from 1 to $d$, $q_{N}, \tau_{0, N}, \tau_{1, N}$ are constants that depend on sample size $N$, and IG $\left(\alpha_{1}, \alpha_{2}\right)$ is the Inverse Gamma distribution with shape parameter $\alpha_{1}$ and scale parameter $\alpha_{2}$.
Under regular conditions (See conditions 4.1–4.5 in~\cite{narisetty2014bayesian}), selection consistency is established:
\begin{theorem}
Assume regular conditions hold, under the model with inverse Gamma prior, we have $p\left(\rz=t \mid \boldsymbol{Y}, \sigma^{2}\right) \stackrel{\mathrm{p}}{\longrightarrow} 1$ as $n \rightarrow \infty$, that is, the posterior probability of the true model goes to $1$ as the sample size increases to $\infty$.
\end{theorem}
More related works on Bayesian feature selection can be found in~\cite{george1997approaches,mitchell1988bayesian}.

\subsection{Simulation Study}\label{FS:sec6}
In this section, we will conduct a series of simulations to verify selection performance on an $i.i.d.$ dataset with varying sizes and dimensions.
Here, we consider cases in the standard multivariate regression. We first generate each input predictor from a standard normal distribution: $x_{n i} \sim N(0,1)$ for $i=1,...,d$, and thus we generate response variables by subsequently sampling $\beta_{j} \sim \operatorname{Uniform}(1,3)$ for $j=1,...,k<d$ and $y_{n} \sim N(\sum_{i=1}^{k}\beta_{i}x_{n i},1)$. We then vary the values of $d$ and $k$ to find the potential influence in terms of True Positive Rate (TPR) and False Positive Rate (FPR). The results are shown in Table~\ref{tab:simulation_result}.

We repeat each case 50 times and present the mean and variance of TPR and FPR. The hyper-parameter setting is listed in Table~\ref{tab:simulation_hp}. We vary $n^{s}$ to study the influence of batch size. Overall, our method illustrates the experimental selection consistency. When $n>d$, our method almost always selects the correct $k$ variables with TPR close to $100\%$ and successfully screens all unnecessary variables with FPR equal to $0\%$. Even under the less informative circumstance when $n$ has an equal or less amount than $d$, our method can still achieve great selection results with TPR above $90\%$. As $n$ goes up,  there is a uniform improvement in all cases in terms of TPR and FPR.

\begin{table}[h]
    \centering
    \caption{Hyper-parameters setting in feature selection.}
    \label{tab:simulation_hp}
    \begin{tabular}{cccccccccc}
        \toprule
        {$\pi_{i}$}&{$\nu_{0,1}$}&{$\nu_{0,2}$}&{$\nu_{i,1}$}&{$\nu_{i,2}$}&{$\nu_{i,3}$}&{$\nu_{i,4}$}&{$T$}&{$n^{s}$}&{$\epsilon$}\\
        \midrule
        $0.5$&$1$&$1$&$1$&$1$&$5$&$1$&$1000$&$256$&$0.5$\\
        \bottomrule
    \end{tabular}
\end{table}

\begin{table}[htbp]
    \centering
    \caption{Feature selection in terms of TPR/FPR.}
    \label{tab:simulation_result}
    \begin{tabular}{cccccccc|ccc}
        \toprule
        \textbf{d=100}&{}&k&{}&n&{}&$n^{s}$&{}&TPR&{}&FPR\\
        \hline
        Case 1 &{}&50&{}&200&{}&64&{}&99.92\%$\pm$0.39\%&{}&0.00\%$\pm$0.00\%\\
        Case 2 &{}&50&{}&200&{}&128&{}& 99.92\% $\pm$ 0.39\%&{}&0.00\%$\pm$0.00\% \\
        Case 3 &{}&50&{}&400&{}&64&{}& 100.00\% $\pm$ 0.00\%&{}&0.00\%$\pm$0.00\% \\
        Case 4 &{}&50&{}&400&{}&128&{}& 100.00\% $\pm$ 0.00\%&{}&0.00\%$\pm$0.00\% \\
        Case 5 &{}&90&{}&200&{}&64&{}&99.86\%$\pm$0.42\%&{}&0.00\%$\pm$0.00\%\\
        Case 6 &{}&90&{}&200&{}&128&{}& 99.93\% $\pm$ 0.26\%&{}&0.00\% $\pm$ 0.00\% \\
        Case 7 &{}&90&{}&400&{}&64&{}& 100.00\% $\pm$ 0.00\%&{}&0.00\% $\pm$ 0.00\% \\
        Case 8 &{}&90&{}&400&{}&128&{}& 100.00\% $\pm$ 0.00\%&{}&0.00\% $\pm$ 0.00\% \\
        \hline
        \textbf{d=300}&{}&k&{}&n&{}&$n^{s}$&{}&TPR&{}&FPR\\
        \hline
        Case 1 &{}&100&{}&300&{}&64&{}&95.21\%$\pm$2.22\%&{}&2.16\%$\pm$1.52\%\\
        Case 2 &{}&100&{}&300&{}&256&{}& 96.46\% $\pm$ 2.12\%&{}&2.31\% $\pm$ 2.10\% \\
        Case 3 &{}&100&{}&500&{}&64&{}& 99.92\% $\pm$ 0.27\%&{}&0.00\% $\pm$ 0.00\% \\
        Case 4 &{}&100&{}&500&{}&256&{}& 100.00\% $\pm$ 0.00\%&{}&0.00\% $\pm$ 0.00\% \\
        Case 5 &{}&250&{}&300&{}&64&{}&91.34\%$\pm$2.92\%&{}&11.92\%$\pm$6.79\%\\
        Case 6 &{}&250&{}&300&{}&256&{}& 91.95\% $\pm$ 2.40\%&{}&14.56\% $\pm$ 8.35\%\\
        Case 7 &{}&250&{}&500&{}&64&{}& 99.92\% $\pm$ 0.17\%&{}&0.00\% $\pm$ 0.00\% \\
        Case 8 &{}&250&{}&500&{}&256&{}& 99.92\% $\pm$ 0.05\%&{}&0.00\% $\pm$ 0.00\% \\
        \hline
        \textbf{d=500}&{}&k&{}&n&{}&$n^{s}$&{}&TPR&{}&FPR\\
        \hline
        Case 1 &{}&100&{}&450&{}&64&{}&92.70\%$\pm$2.56\%&{}&4.41\%$\pm$1.67\%\\
        Case 2 &{}&100&{}&450&{}&256&{}& 92.89\% $\pm$ 2.69\%&{}&4.90\% $\pm$ 1.82\% \\
        Case 3 &{}&100&{}&800&{}&64&{}& 99.94\% $\pm$ 0.23\%&{}&0.00\% $\pm$ 0.00\% \\
        Case 4 &{}&100&{}&800&{}&512&{}& 100.00\% $\pm$ 0.00\%&{}&0.00\% $\pm$ 0.00\% \\
        Case 5 &{}&450&{}&500&{}&64&{}&90.21\%$\pm$2.56\%&{}&12.68\%$\pm$6.38\%\\
        Case 6 &{}&450&{}&500&{}&256&{}& 92.06\% $\pm$ 1.84\%&{}&16.04\% $\pm$ 6.69\%\\
        Case 7 &{}&450&{}&800&{}&64&{}& 99.92\% $\pm$ 0.13\%&{}&0.00\% $\pm$ 0.00\% \\
        Case 8 &{}&450&{}&800&{}&512&{}& 100.00\% $\pm$ 0.00\%&{}&0.00\% $\pm$ 0.00\% \\
        \bottomrule
    \end{tabular}
\end{table}